\documentclass[journal, twocolumn]{IEEEtran}

\IEEEoverridecommandlockouts
\usepackage{graphicx}
\usepackage{graphics}
\usepackage{xcolor}
\usepackage[skip=0pt]{subcaption}
\usepackage{amsmath}
\usepackage{gensymb}
\usepackage{amssymb}
\usepackage{algorithm}
\usepackage{algorithmicx}
\usepackage[noend]{algpseudocode}
\usepackage{booktabs}
\usepackage{titlesec}
\usepackage{listings}
\usepackage{enumitem}
\usepackage{multirow}
\usepackage{placeins}
\usepackage{rotating}
\usepackage{pifont}
\usepackage{array}
\usepackage{mdframed}
\usepackage{lipsum}
\usepackage[utf8]{inputenc}
\usepackage[colorlinks=true, allcolors=blue]{hyperref}
\usepackage[many]{tcolorbox}
\usepackage{varwidth}
\usepackage{cleveref}
\usepackage{colortbl}
\usepackage{array}
\usepackage{blindtext}
\usepackage{makecell}

\usepackage{subcaption}

\title{\LARGE Waypoint Planning Networks\vspace{-2 mm}}

\author{
\vspace{-0 mm}
Alexandru-Iosif Toma$^{\dagger}$, Hussein Ali Jaafar$^{\star}$, Hao-Ya Hsueh$^{\star}$, Stephen James$^{\dagger}$, Daniel Lenton$^{\dagger}$,\\
Ronald Clark$^{\dagger}$,  Sajad Saeedi$^{\star}$% <-this % stops a space
\vspace{-5 mm}
\thanks{$^{\dagger}$Imperial College London\quad\quad $^{\star}$Ryerson University}%
}

\begin{document}
\maketitle
\vspace{-2 mm}

\begin{abstract}

With the recent advances in machine learning, path planning algorithms are also evolving; however, the learned path planning algorithms often have difficulty competing with success rates of classic algorithms. We propose \emph{waypoint planning networks (WPN)}, a hybrid algorithm based on LSTMs with a local kernel---a classic algorithm such as A*, and a global kernel using a learned algorithm. 
WPN produces a more computationally efficient and robust solution. We compare WPN against A*, as well as related works including motion planning networks (MPNet) and value iteration networks (VIN). In this paper, the design and experiments have been conducted for 2D environments. 
Experimental results outline the benefits of WPN, both in efficiency and generalization. It is shown that WPN's search space is considerably less than A*, while being able to generate near optimal results. Additionally, WPN works on partial maps, unlike A* which needs the full map in advance. The code is available online\footnote{\href{https://sites.google.com/view/waypoint-planning-networks/home}{https://sites.google.com/view/waypoint-planning-networks}}.

\end{abstract}

%===============================================================================

\section{Introduction}
Nowadays, mobile autonomous systems are indispensable, used to automate tasks that cannot be performed at large scale or in a safe manner by a human being. Some example applications include search-and-rescue, large-scale manufacturing, and warehouse management. 
Path planning algorithms have essential applications in these mobile autonomous systems, motivating researchers to design efficient algorithms \cite{gonzalez2016review}.

Currently, there are many classic solutions to the path planning problem. Popular algorithms include A* \cite{choset2005principles, duchovn2014path, zhang2014multiple, 5937169}, Rapidly-exploring Random Tree (RRT) family \cite{lavalle1998rapidly, rodriguez2006obstacle, lavalle2001randomized, karaman2011sampling}, and value iteration on Markovian decision processes (MDP) \cite{szepesvari2010algorithms, satia1973markovian}. Furthermore, these algorithms are often offline, requiring complete knowledge of the environment beforehand. Advancements in learned path planning algorithms have been rapid; however, they are often unable to compete with the success rates of classic algorithms.
\textcolor{black}{Search or graph-based algorithms such as A* and Dijkstra produce optimal solutions but require to search a large portion of the configuration space. This becomes computationally expensive in large maps or motions with high degrees of freedom. Though sampling-based algorithms such as RRT reduce the space/time complexity, the results are often sub-optimal. The problem that we try to tackle in this paper is to reduce the search space, and therefore the time/space complexity, of the A* algorithm by learning better heuristics.}

\begin{figure}[t!]
  \centering
  \begin{subfigure}[b]{0.48\linewidth}

    \includegraphics[width=\linewidth]{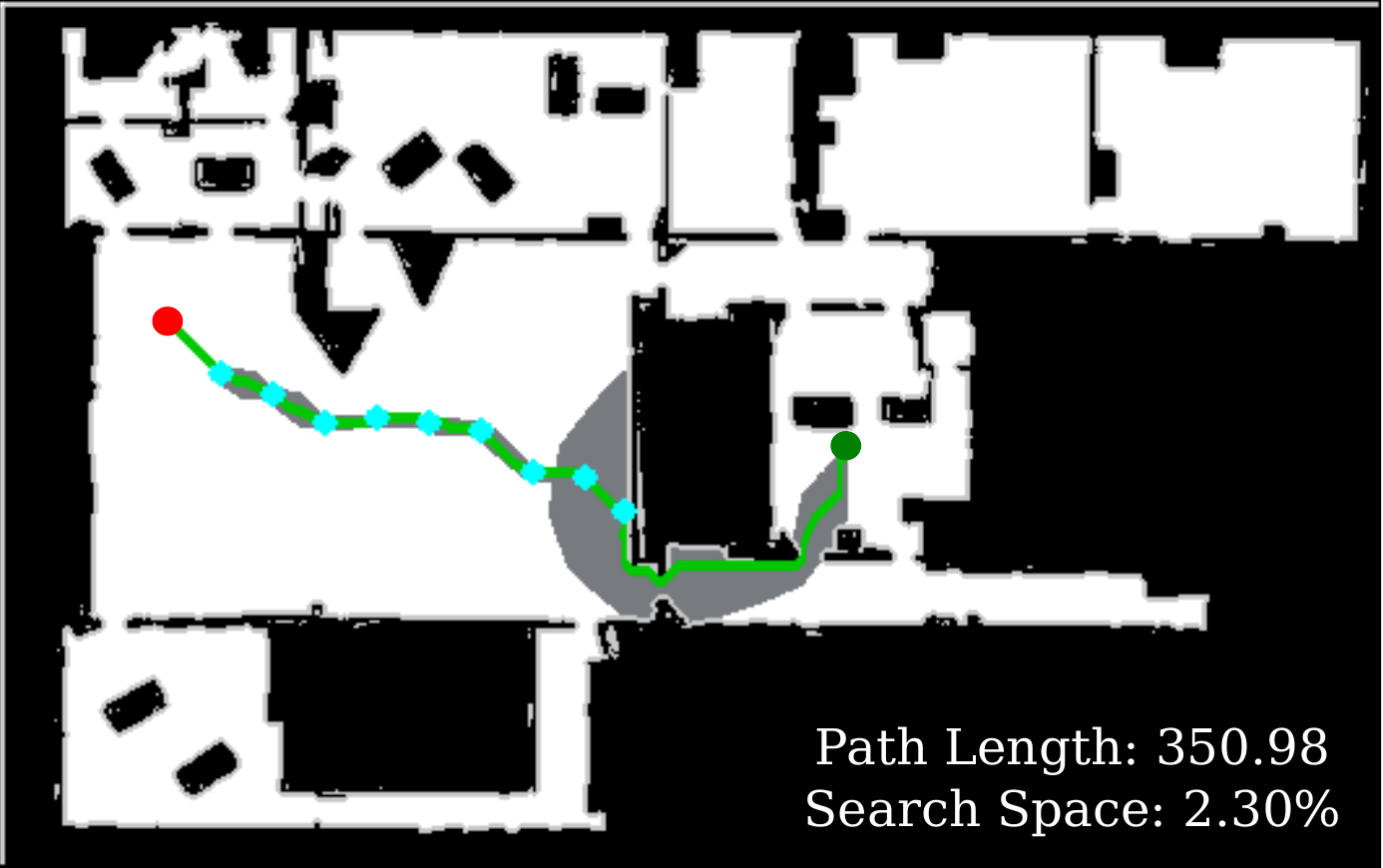}
    \caption*{\centering \scriptsize (a) waypoint planning networks (WPN)}
  \end{subfigure}
  \begin{subfigure}[b]{0.48\linewidth}

    \includegraphics[width=\linewidth]{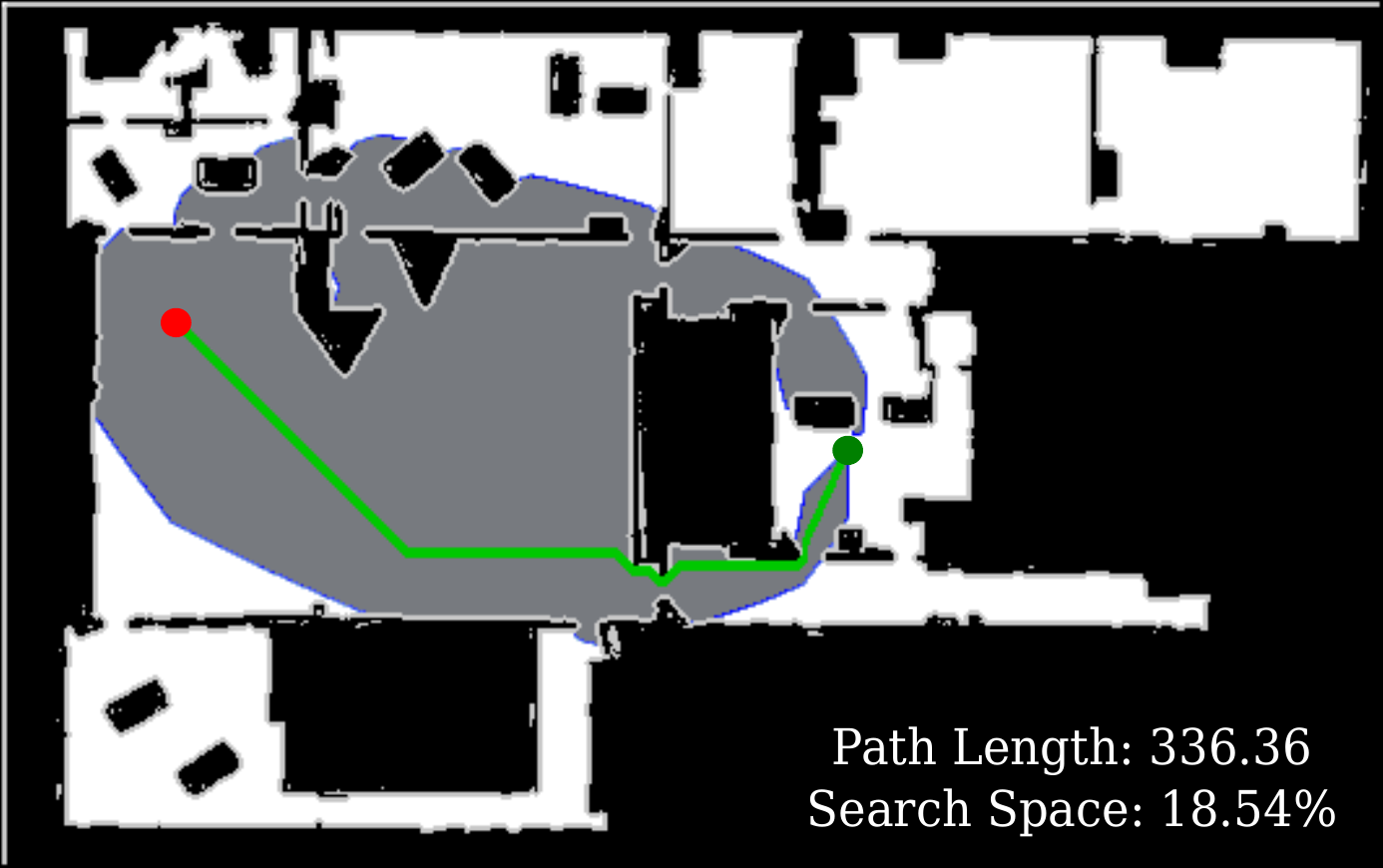}
    \caption*{\scriptsize (b) A* algorithm}
  \end{subfigure}

    \caption{Evaluation of waypoint planning networks and A* paths on a real map, from the red start point to the goal.
    Gray cells show the search space. Cyan disks show the waypoints. WPN's search space is significantly smaller than A*.} 
    \vspace{-5 mm}
  \label{fig: intro_show}
\end{figure}

To solve these issues, we propose \textit{waypoint planning networks (WPN)}, a hybrid algorithm using a local kernel, typically a classic algorithm such as A*, and a global kernel using a learned algorithm. 
In this paper, the scope of the design and experimentation is limited to 2D planning, i.e. floorplan-like environments, which is suitable for many indoor mobile robotic applications.
We show that WPN produces a more computationally efficient and robust solution. It is able to work with partial knowledge of the environment. To evaluate WPN, we generate maps with a benchmarking platform, PathBench \cite{pathbench}, and compare it against A*, as well as related works including value iteration networks (VIN)~\cite{tamar2016value} and motion planning networks (MPNet) \cite{qureshi2019motion}. 
Experimental results outline the benefits of WPN, both in efficiency and generalization. WPN generates near-optimal solutions but with reduced search space, compared with A*. \textcolor{black}{A reduced search-space, which directly translates to reduced memory utilization, is particularly important in robotics and embedded devices with limited resources.} Fig.~\ref{fig: intro_show} shows an example run of the algorithm compared with A*. In this case, the search space, i.e. the visited cells, by both algorithms are shown in gray.

\begin{figure*}[ht!]
    \centering
    \includegraphics[width=.95\textwidth]{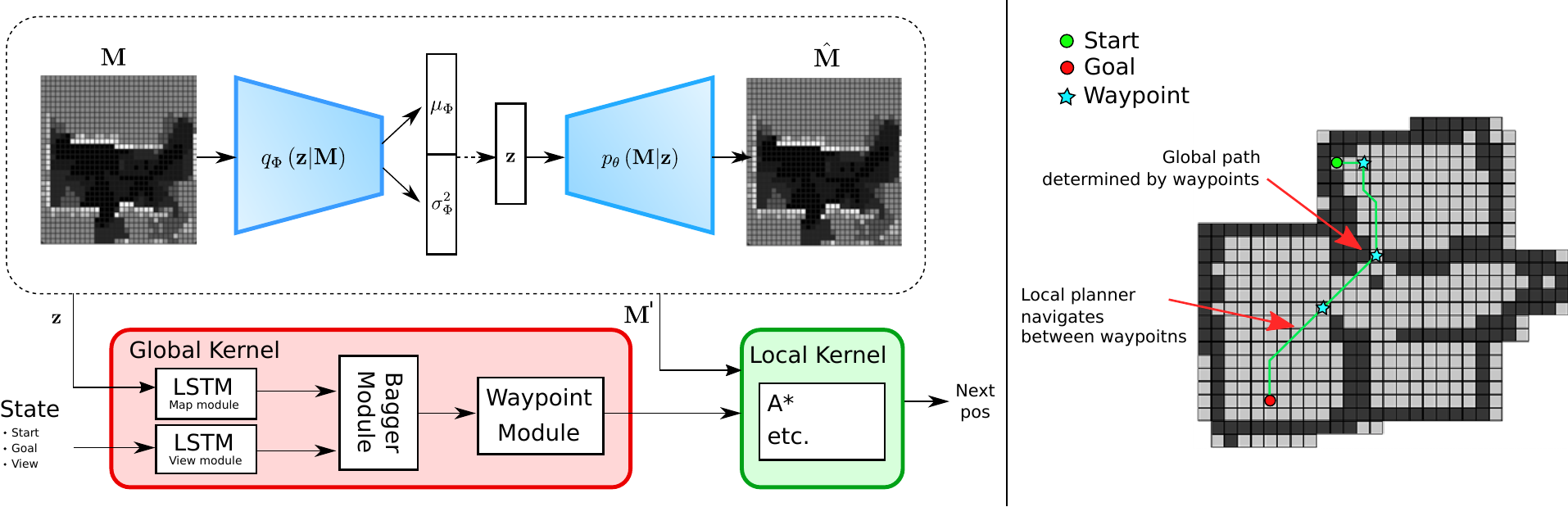}
    \caption{Overview of the waypoint planning networks (WPN) architecture. \textcolor{black}{The start point, goal point, and a compact representation of the map are used to generate waypoints between start and goal points. A*, as a local kernel, is used to plan paths between the waypoints.} }
    \label{fig:my_label}
        \vspace{-5 mm}
\end{figure*}

\section{Related Work}

\textbf{Classic planning} methods can be broadly categorized into four groups: graph search, sampling-based, interpolating curve, and numerical optimization \cite{gonzalez2016review}. Graph search performs its search in a state space that is generally represented as an occupancy grid or lattice, and encompasses  prominent algorithms such as A* \cite{choset2005principles, duchovn2014path, zhang2014multiple, 5937169}, Dijkstra \cite{choset2005principles, zhang2014multiple}, wavefront \cite{choset2005principles, luo2014effective}, and bug algorithms \cite{choset2005principles, rajko2001pursuit}, to name a few. Unlike graph search, sampling-based planners, e.g. Rapidly-exploring Random Tree (RRT) \cite{lavalle1998rapidly} and Probabilistic Roadmap (PRM) \cite{kavraki1994probabilistic}, can operate in high dimensional spaces more efficiently by randomly sampling in either the configuration or state space, but these methods come at a cost of sub-optimal solutions. Where the previous two methods plan on the global scale, interpolating curve planners \cite{reeds1990optimal, funke2012up, xu2012real} use local planning; given a set of waypoints, these methods generate a local path whilst optionally considering a set of constraints. Finally, numerical optimization methods, aim to optimize an objective function, often with a set of constraints. Previously, these methods were often used to smooth prior computed trajectories \cite{dolgov2010path} or used to compute trajectories from kinematic constraints \cite{ziegler2014making}; however, with the advent of deep learning, these optimization methods have become very diverse, to which we now discuss.

\textbf{Learning-based planning algorithms} have increasingly become more common \cite{Chen2016Humanoids}, 
\cite{gupta2017cognitive}, 
\cite{inoue2019robot}, 
\cite{NeuralRRT}. 
In most learning-based path planning algorithms, 
imitation learning~\cite{ross2011reduction} plays a key role~\cite{TDPPNet}. 
Neural networks have been used to improve the classic algorithms, for instance by adaptively sampling a particular region of a configuration space in sampling-based algorithms~\cite{qureshi2018deeply}. A similar concept was used by Chamzas et al. \cite{chamzas2019using}. In both works the computational complexity has been reduced compared to classic methods. 
Other methods use neural networks to generate a complete path. For instance, motion planning networks (MPNet) encodes the point cloud measurements of the workspace to generate a path from start to goal~\cite{qureshi2019motion}, \cite{qureshi2020motion}. MPNet works 3$\times$ faster than BIT* \cite{Bit*}. 
Qureshi et al. formulate constraints into MPNet, and present CoMPNet, which encompasses kinematic constraints \cite{CoMPNet}. A similar work is done in~\cite{Johnson_IROS_2020}.
Bency et al. present OracleNet, a recurrent neural network (RNN)-based approach to generate fast near-optimal paths for robotic arms~\cite{bency2019neural}. OracleNet needs training on each new environment which makes the algorithm suitable for static environments. 

\textbf{Reinforcement Learning} (RL) approaches such as value iteration networks (VINs)~\cite{tamar2016value},  \cite{Levine2013},  learning-from-demonstration (LfD)~\cite{Abbeel2010}, guided policy search (GPS)~\cite{Levine2013}, and universal planning networks (UPN)~\cite{srinivas2018universal} have also been used for path planning. Wu et al. present three-dimensional path planning network (TDPP-Net) \cite{TDPPNet}, which is an end-to-end network that predicts 3D actions via 2D CNNs. TDPP-Net learns a policy via supervised imitation learning from the Dijkstra’s algorithm. These methods often use differentiable models of neural networks and are able to learn to plan. We compare WPN with VINs, as well as MPNet~\cite{qureshi2019motion} and two other RNN-based algorithms \cite{nicola2018lstm} and \cite{inoue2019robot}. The comparisons are preformed on synthetic maps of different sizes~\cite{pathbench}, real-world maps~\cite{first_map}, \cite{second_map}, \cite{third_map}, and HouseExpo maps \cite{houseexpo}.

\section{Network Architecture} \label{sec:wpn} 

Fig. \ref{fig:my_label} demonstrates the architecture of the waypoint planning networks. The inputs of the network are the current (partial) 2D map, the start point, and goal point. Given the start point, WPN finds a waypoint and repeats the process from the waypoint as the next start point, until it achieves the goal. The algorithm has a global kernel and a local kernel. The global kernel is responsible for finding a set of waypoints towards the goal. The local kernel uses a classic algorithm to find a path between the waypoints. The global kernel has four modules: 
\emph{View Module},
\emph{Map Module}, 
\emph{Bagging Module}, and
\emph{Waypoint Module}. 

The view module utilizes the current information provided by the sensor, in this case a typical measurement from a scanning laser ranger. The map module utilizes the current (partial) map of the environment, in this case a grid map. The bagging and waypoint modules together find the best waypoints. These modules are explained next. 

\subsection{View Module} \label{sec:view_module}

The \emph{view module} uses an LSTM network \cite{hochreiter1997long} to retrieve the next action that the agent should take given the current location data (pose and local surroundings information). This module is based on \cite{nicola2018lstm}, with some architectural and logic changes. The concept behind having this module is to allow the network to learn a path towards a goal in simple environments. 
The LSTM architecture takes four inputs: (1) the normalized distance between the agent to obstacles on all eight directions of the Moore neighborhood, \textcolor{black}{an 8D vector}, 
(2) the normalized direction to the goal, \textcolor{black}{a 2D vector,}  
(3) the angle defined by the direction to the goal (not required to be normalized as it is already bounded by definition), and 
(4) the normalized distance to the goal. 
The model contains a hidden state and cell state which are initialized at each new batch with a zero-tensor of size $2 \times lstm\_layers \times batch\_size \times lstm\_output\_size$. The architecture has the following structure: one batch normalization layer, two LSTM layers, one batch normalization layer, and one linear layer. The network uses the cross entropy loss function. 
This module exhibits a greedy behaviour. It has a good success rate when there are direct routes, but fails when the path is complex, such as u-turns.

\subsection{Map Module} \label{sec:map_module}

The \emph{map module} has both Convolutional Auto-encoder (CAE)~\textcolor{black}{~\cite{inoue2019robot}, similar to \cite{Everett_IROS_2019},} and LSTM components and attempts to fix some issues, e.g. greediness, that are present in the \emph{view module} (i.e. the algorithm does not know how to navigate between complex obstacles and long corridors). This is done by augmenting the input from the LSTM network with the compressed global image snapshot. When we are dealing with partially known environments, we still use the global image snapshot, but we include the unknown environment as well. The global snapshot is compressed using a CAE (See Fig.~\ref{fig:my_label}, top left, \textcolor{black}{$M$ is the map, $\hat{M}$ is the reconstructed map, and $z$ is the learned latent variable}).
The CAE encoder contains four convolutional layers and one linear layer. 
Each convolution layer is composed of multiple layers placed in the following order: convolutional layer, batch normalization layer, max pool, and leaky ReLU as the activation function. The final layer of the encoder is a linear layer with another batch normalization layer. 
The CAE decoder contains one linear layer and four de-convolutional layers.  Each de-convolutional layer is composed of multiple layers placed in the following order: de-convolutional layer, batch normalization layer, and ReLU activation function. The last de-convolutional layer has Tanh activation function as the input is normalized in the range [-1, 1] and the output of Tanh matches it.
The CAE is trained on the map training datasets, which is a collection of synthetically generated maps. 

\subsection{Bagging Module} \label{sec:bagging}
The \emph{view} and \emph{map modules} behave differently depending on the map layout. Moreover, the same behaviour variability exists when training the models on uncorrelated datasets.
The \emph{bagging module} 
is a solution inspired by ML ensemble methods which combines the previous 
two modules into a unique best-of-all kind of algorithm~\cite{dietterich2000ensemble}. 
The bagging planner attempts to boost the performance of the previous modules by picking the best solution depending on the layout of the environment.

Ensemble ML methods use multiple weak learners and output a majority voting consensus. Ensemble ML is split into two categories: sequential and parallel. Sequential ensemble ML (e.g. AdaBoost) trains the weak learners sequentially on the same dataset.
We focus on the parallel ensemble methods which train all weak learners at the same time in parallel, but on different training datasets sampled from the original dataset. Thus, the weak learners are not correlated, and each one of them learns different features. Lastly, by having multiple uncorrelated weak learners, the voting procedure increases the accuracy of the predictions. 

We use the \emph{view} and \emph{map} modules as weak learners. Since the models can be trained on different training datasets, weak learners learns how to behave in different environments and are uncorrelated. In run time, weak learners are executed in parallel on the map. If any/multiple kernels have found the goal, we pick the one which has lower traversed length. Otherwise, we pick the kernel which has made the furthest progress. Using the \emph{bagging} module, the success rate of finding the goal is significantly increased compared to the \emph{view} and \emph{map} modules. 

\subsection{Waypoint Module}
The previous three modules, \emph{view}, \emph{map}, and \emph{bagging} are able to generate paths from start to goal points sequentially. However, the success rates of these modules, as shown in the experiments, are still not comparable to the success rate of A*. This is particularly more evident in long sequences. The \emph{waypoint} module has been designed to fix this problem and increase the success rate.

The \emph{waypoint} module is responsible for suggesting a series of waypoints which will guide the agent through the environment. To generate waypoints, any of the previous three modules may be use, but the \emph{bagging} module performs the best. The algorithms used for waypoint generation is referred to as a global algorithm (kernel), GK, for simplicity. The waypoint generation is achieved by bounding the number of iterations of the global kernel, e.g. the bagging module.

Once the waypoints are known, a local kernel, LK, is responsible for planning a path for actual manoeuvring between the waypoints. Any classic solution can be used as the local planner, but we have decided to use A* as it represents the base algorithmic frame of reference against all other proposed solutions. \textcolor{black}{The waypoint module is essential in the achieved success rates of WPN. This is demonstrated in experiments.}

\section{Experimental Results}
\label{sec:result}

\begin{figure}[]
  \begin{subfigure}[b]{0.01\linewidth}
  \centering
    \caption*{}
    \end{subfigure}
    \hfill
  \begin{subfigure}[b]{0.31\linewidth}
  \caption*{\scriptsize Uniform Random-fill}
  \end{subfigure}
  \hfill
  \begin{subfigure}[b]{0.31\linewidth}
  \caption*{\scriptsize Block}
  \end{subfigure}
  \hfill
  \begin{subfigure}[b]{0.31\linewidth}
  \caption*{\scriptsize House-style}
  \end{subfigure}
  \hfill
  \vspace{-.5 mm}    
  \begin{subfigure}[b]{0.01\linewidth}
  \centering
    \caption*{\rotatebox{90}{\scriptsize \quad \quad \quad \quad  \, WPN}}
    \end{subfigure}
    \hfill
  \begin{subfigure}[b]{0.31\linewidth}
    \includegraphics[width=\linewidth]{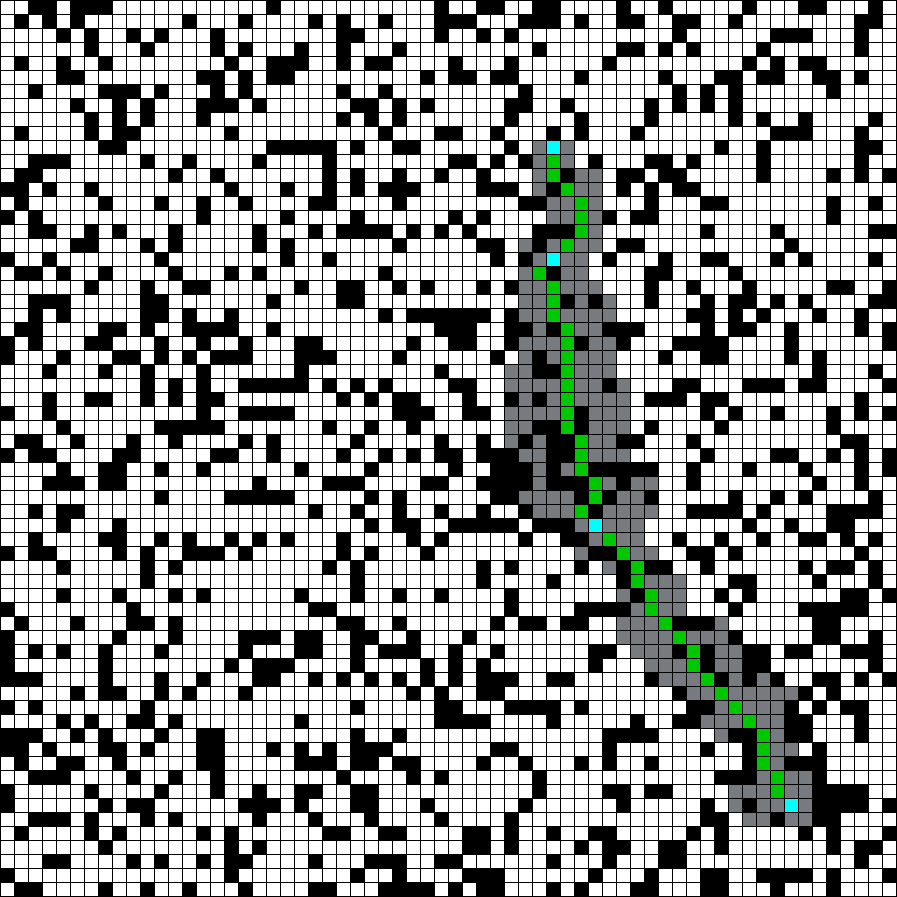}
     \caption{\scriptsize 57.36 (m), \textbf{185} (cell)}

  \end{subfigure}
  \hfill
  \begin{subfigure}[b]{0.31\linewidth}
    \includegraphics[width=\linewidth]{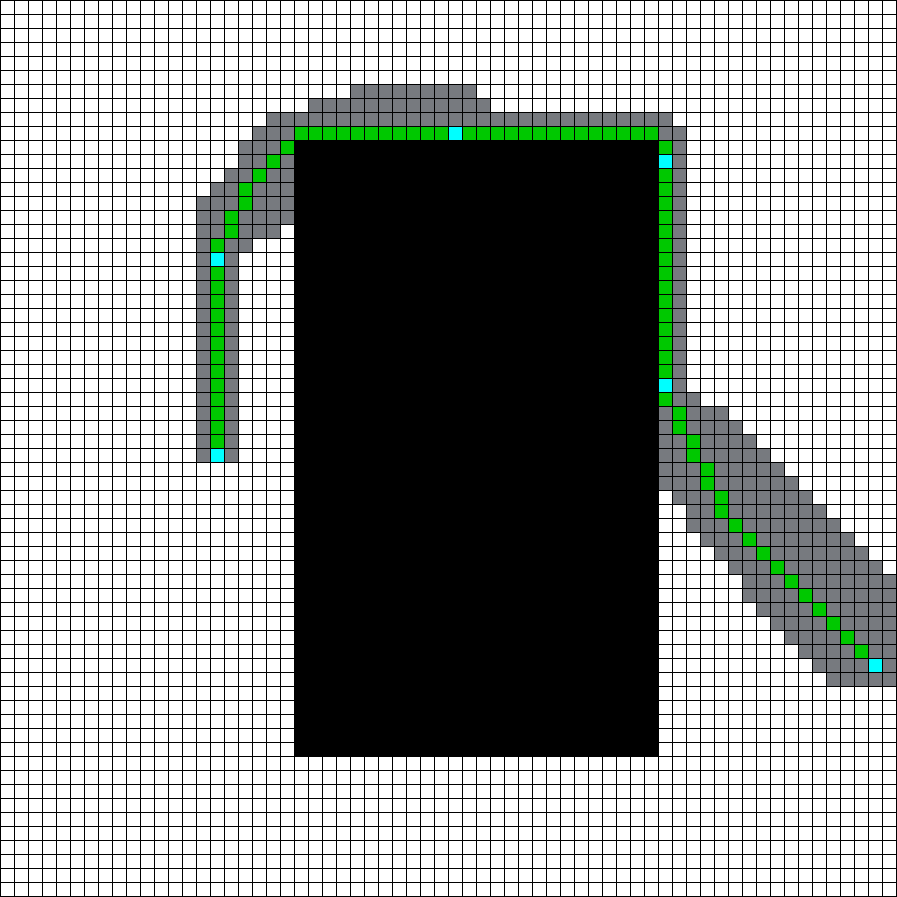}
     \caption{\scriptsize 95.11 (m), \textbf{387} (cell)}
     
  \end{subfigure}
  \hfill
  \begin{subfigure}[b]{0.31\linewidth}
    \includegraphics[width=\linewidth]{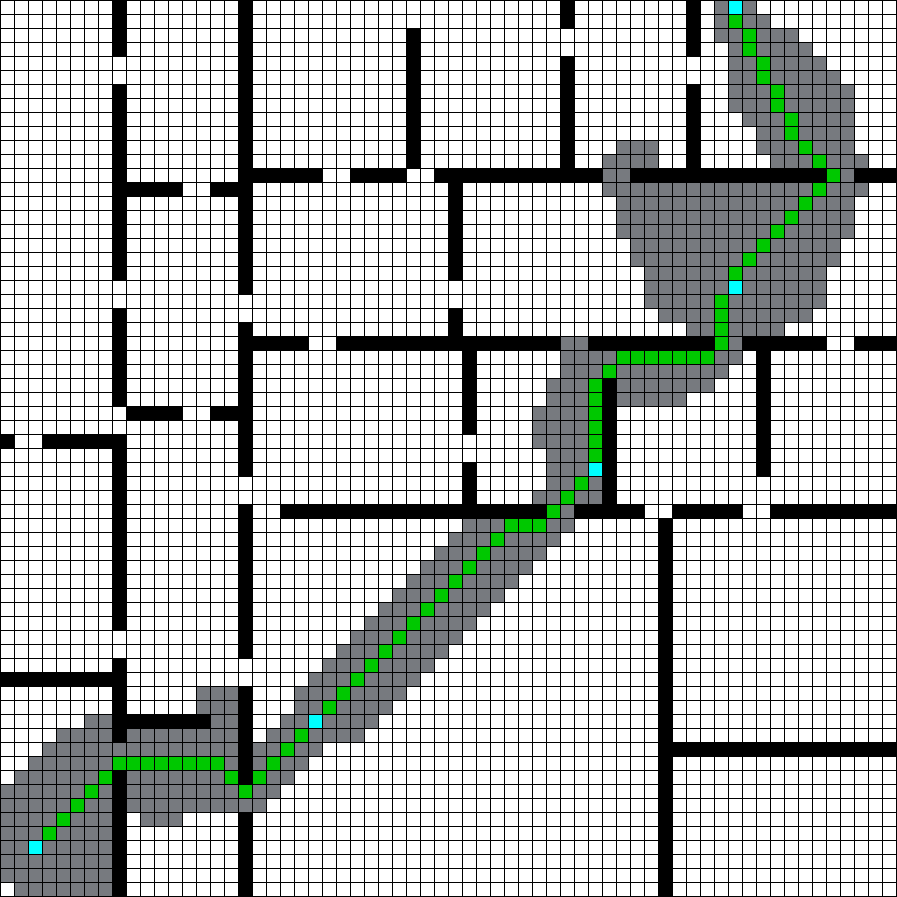}
     \caption{\scriptsize 99.30 (m), \textbf{618} (cell)}
  \end{subfigure}
  \label{fig: Way runs}

  \vspace{.5 mm}    
  \begin{subfigure}[b]{0.01\linewidth}
  \centering
    \caption*{\rotatebox{90}{ \scriptsize \quad \quad \quad \quad  \quad  \, A*}}
    \end{subfigure}
  \hfill
  \begin{subfigure}[b]{0.31\linewidth}
    \includegraphics[width=\linewidth]{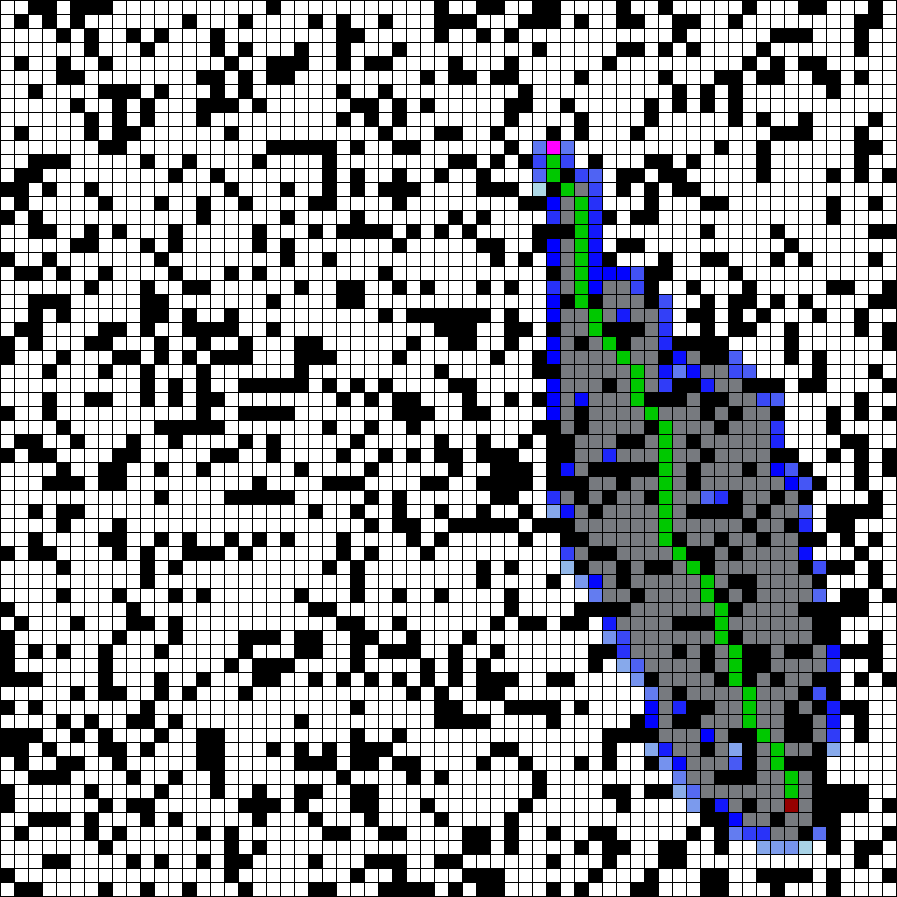}
    \caption{\scriptsize \textbf{54.04} (m), 467 (cell)}
  \end{subfigure}
  \hfill
  \begin{subfigure}[b]{0.31\linewidth}
    \includegraphics[width=\linewidth]{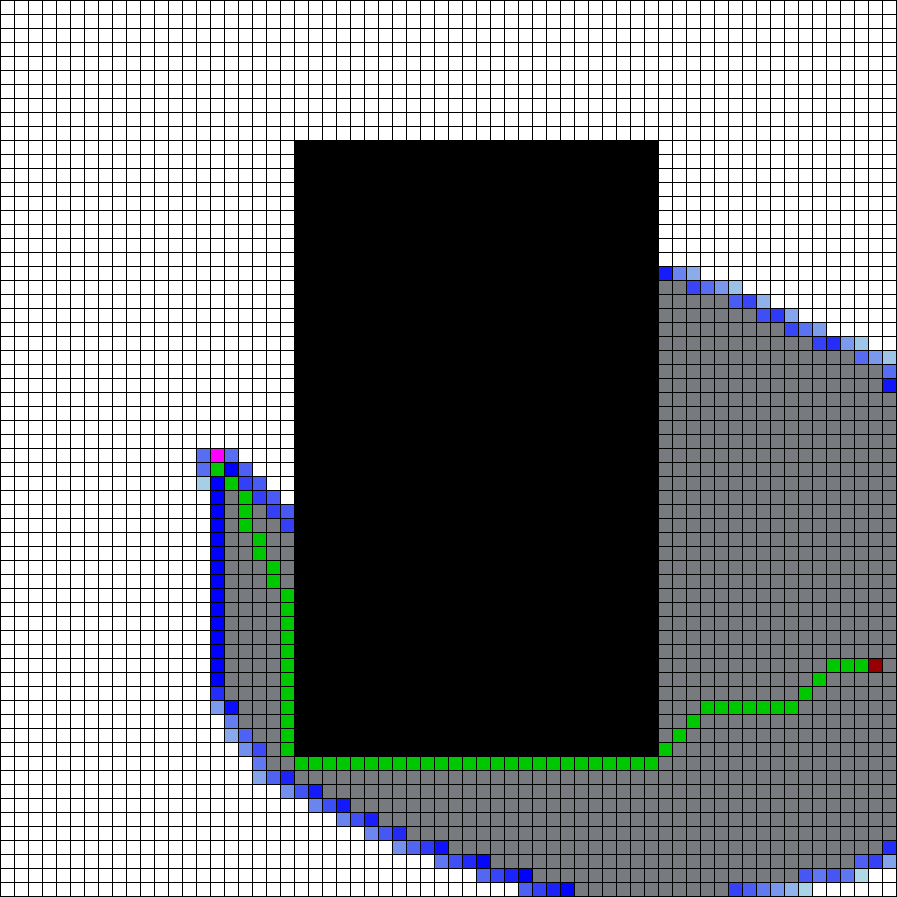}
    \caption{\scriptsize \textbf{68.38} (m), 1054 (cell)}
  \end{subfigure}
  \hfill
  \begin{subfigure}[b]{0.31\linewidth}
    \includegraphics[width=\linewidth]{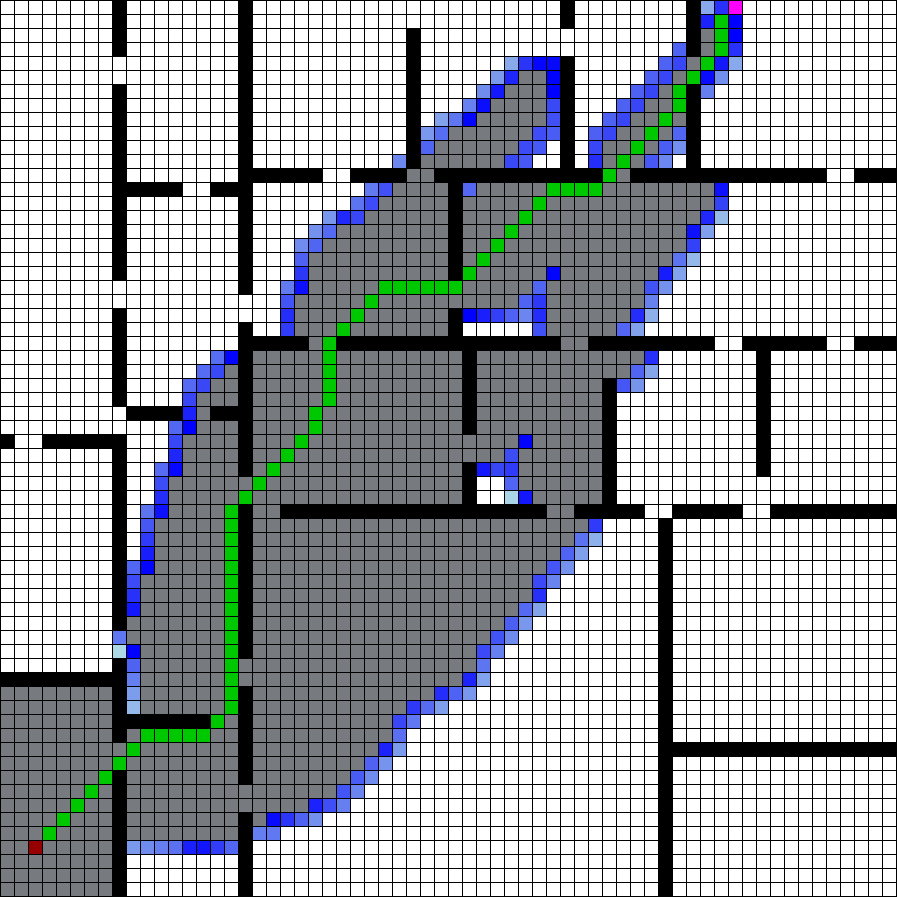}
    \caption{\scriptsize \textbf{87.74} (m), 1381 (cell)}
  \end{subfigure}
    \vspace{.5 mm} 
  \caption{WPN (first row) vs A* (second row) in random, block, and house maps. Gray cells show the search space. Numbers indicate \emph{path length (m) and search space (cell)}.}
  \label{fig: Way runs comp}
\end{figure}

\begin{table*}[t]
    \centering

\caption{Generalized benchmarking test results. All algorithms except VIN were trained on 30,000 maps of size 64$\times$64. They were then tested on 3000 of various sized unseen maps. VIN was trained differently, for each map size, a model was trained only with the same size maps. Metrics are defined in the text.}    
    \resizebox{18cm}{!}{
    \begin{tabular}{ll|ccc|ccc|ccc|c}
        \toprule
        Size & Metric  & WPN & WPN-view & WPN-map & A* & VIN~\cite{tamar2016value} & VIN64  & Bagging Mod. & View Mod.~\cite{nicola2018lstm} & Map Mod.~\cite{inoue2019robot} & MPNet ~\cite{qureshi2019motion} \\
        \midrule

        \multirow{3}{*}{8$\times$8} &
        Succ. R \%  & \textbf{91.60}  & \textbf{91.60} &   \textbf{91.60} & \textbf{91.60}  &  90.26 & 71.06  & 87.70  & 80.17  & 70.60 & 41.4\\
        & Dev. \%   & -3.54  &  -8.33 &  -7.29 &  \textbf{0} & -21  & -27.6  & -3.04  & -3.38 & -15.88 & -12.1  \\
        & Comp. (sec)   & 0.2045  &  0.0197 & 0.0457 & \textbf{0.0009} & 0.0118  & 0.0094  &  0.2289 & 0.0166 & 0.0387 & 0.1894 \\
        & Dist. Left & 0.32  & \textbf{0.31}  & 0.35  & 0.43 & 3.37  & 3.89  & 0.45  & 0.70 & 1.00 & 2.72 \\
        & Search \% & 13.79  & 13.65  & \textbf{12.50}  & 26.37 & NA  & NA  & NA  & NA & NA & NA \\
        \midrule
        
        \multirow{3}{*}{16$\times$16} 
        & Succ. R \%  & \textbf{97.07}  & \textbf{97.07}  & \textbf{97.07 } & \textbf{97.07} & 92.20  & 71.23  & 89.63  & 76.57 & 64.43 & 33.43 \\
        & Dev. \%   & -6.45  & -11.26  & -9.93  & \textbf{0} & -35  & -39.80  & -5.41  & -5.34 & -16.43 & -38.2 \\
        & Comp. (sec)   & 0.277  & 0.0284  & 0.0633 & \textbf{0.0032} & 0.032  & 0.028  &  0.2359 & 0.0154 & 0.0277 & 0.3385\\
        & Dist. Left  & \textbf{0.19}  & \textbf{0.19 } &  0.23 & 0.31 & 7.73  & 6.208  & 0.61  & 1.33 & 1.33 & 6.07 \\
        & Search \% & 8.61  & 8.26  & \textbf{7.09 } & 17.84 & NA  & NA  & NA  & NA & NA & NA\\
        \midrule
        
        \multirow{3}{*}{28$\times$28} 
        & Succ. R \%  & \textbf{97.07}  & \textbf{97.07}  & \textbf{97.07} & \textbf{97.07} & 79.06  & 58.85  & 83.80  & 66.10 & 52.03 & 29.4 \\
        & Dev. \%   &  -14.07 & -16.65  & -12.87 & \textbf{0} & -56  & -66  & -6.63  & -6.27 & -17.2 & -37.8 \\
        & Comp. (sec) &  0.5706 & 0.0534  & 0.0887 & \textbf{0.0065} & 0.0992 & 0.0683 & 0.3341  & 0.0228 & 0.0371 & 0.4353 \\
        & Dist. Left  & \textbf{0.29} & 0.31  & 0.35 & 0.54 & 21.9  & 12.21  & 1.51  & 3.4 & 4.62 & 11.61 \\
        & Search \% & 5.50  & 5.41  & \textbf{4.67} & 14.57 & NA  & NA & NA & NA & NA & NA\\
        \midrule
        
        \multirow{3}{*}{64$\times$64} & Succ. R \%  & \textbf{98.13}& \textbf{98.13}  & \textbf{98.13} & \textbf{98.13} & 26.20  & 26.20  & 79.67  & 57.33 & 46.13 & 16.27 \\
        & Dev. \%   & -23.6  & -22.0  & -14.6 & \textbf{0} & -81.74  & -81.7  & -7.21  & -8.0 & -7.19 & -36.0\\
        & Comp. (sec)   & 1.667  & 0.200  & 0.2286 & \textbf{0.0327} & 0.235 & 0.235 & 0.7711  & 0.0622  & 0.0747 & 0.9109 \\
        & Dist. Left  & \textbf{0.53}  & 0.55  & 0.61 & 0.79 & 37.2  & 37.2  & 4.09  & 9.69 & 11.63 & 30.30 \\
        & Search \% & \textbf{2.03}  & 2.39  & 2.53 & 10.82 & NA & NA  & NA  & NA & NA & NA\\
        \bottomrule
    \end{tabular}}
    \label{tab:specific}
    \small
\end{table*}

In this section, several experiments are presented, under three subsections, briefly outlined below. The first two experiments compare WPN against bagging module, 
view module~\cite{nicola2018lstm}, 
map module~\cite{inoue2019robot},  
VIN~\cite{tamar2016value}, and MPNet~\cite{qureshi2019motion}.\par

\noindent\emph{\bf\,\,1) Generalized Benchmarking}: This benchmarking uses the models trained on a variety of synthetic maps, generated in PathBench \cite{pathbench}. They are trained on 30,000 64$\times$64 size maps, split equally between uniform random-fill, block, and house-style maps (See Fig.~\ref{fig: Way runs comp}). They are then tested on a test-set of 3000 maps, equally divided by size and type (house and uniform random-fill). This highlights the ability of the algorithm to generalize on map sizes it has not been trained on. \par

\noindent\emph{\bf\,\,2) HouseExpo Benchmarking}: This benchmarking involves the generalized models running on maps from the HouseExpo dataset~\cite{houseexpo}, modified to a 100$\times$100 size. \par

\noindent\emph{\bf\,\,3) Real-world}: We run WPN on a robot in real-world and also demonstrate WPN vs A* on a few real-world maps and a Gazebo world.

\noindent\emph{\bf\,\,4) Large maps}: Finally, we test WPN, A*, View and Map modules on larger environments (512$\times$512). This outlines the importance of the waypoint module.

\subsection{Generalized Benchmarking: Synthetic Maps}
\label{sec:Generalized}
For training of WPN, View Module~\cite{nicola2018lstm}, Map Module~\cite{inoue2019robot} and Bagging module, three types of synthetic maps of size 64$\times$64 pixels were procedurally generated: uniform random-fill map, block map, and house-style map (see Fig.~\ref{fig: Way runs comp} for samples). In these maps, start and goal points are chosen randomly. Evaluations are done over maps that have never been seen by the algorithms. The algorithms were trained on 30,000 64$\times$64 maps, equally divided between the three map types. They were then tested on 3000 of each sized map, 8$\times$8 maps, 16$\times$16, and 28$\times$28  maps, split between uniform random-fill and house maps. They were also tested on 1500 64$\times$64 maps, of the same format. \textcolor{black}{Training parameters are as follows: view module was trained with 100 epochs, batch size of 50, input size of 12, and output size of 8 Map module was trained with 50 epochs, batch size of 50, input size of 112, and output size of 8. Training logs can be found in the GitHub repo.}

To demonstrate why such an architecture was chosen for WPN, we present the performance of its individual modules, i.e. \emph{view}, \emph{map}, \emph{bagging} modules. Note that these do not produce any waypoints, rather they plan a path in one go, cell by cell. As discussed in the previous section, the bagging module is the best of map and view modules and that is why it outperforms the other two.

In addition to the WPN algorithm, we will include two other variation of WPN. These are different only in their global kernels, GK. \emph{WPN} uses the bagging module as GK, while \emph{WPN-view} uses the view module and \emph{WPN-map} uses the map module as GKs.

We compare WPN and all its variants with VIN~\cite{tamar2016value}. VIN was trained four times to produce four different models. One model for each size, 8$\times$8 maps, 16$\times$16 maps, and 28$\times$28 maps are trained on 60,000 maps. The 64$\times$64 model was trained on 30,000 maps due to computational limitations. The testing uses the corresponding trained model. This approach of training gives VIN a competitive advantage; each models is tuned for a specific size.
In addition to the standard implementation of VIN based on \cite{tamar2016value}, we also compare the results with VIN64. VIN64 was only trained on 30,000 64$\times$64, like WPNs, then tested on all sizes, similar to WPNs.

MPNet was trained in the same manner as VIN64. MPNet uses two neural models for planning, with the first being an encoder network that embeds obstacle point clouds into latent space and the second network being the planning network that learns to plan a path with the map embedding~\cite{qureshi2019motion}. MPNet's encoder and planning networks were trained similar to the approach taken for the training of the WPN variations.

To evaluate the results maps, five metrics are used:
\begin{enumerate}
    \item {\bf Succ. R}: success rate, defined as the percentage of the successful trajectories created from the entire test. Note that in VIN, the maps with no paths between the start and goal points were discarded, leading to 100\% success rate. We define the success rate differently, by not discarding the maps when no paths exist, to calculate the \emph{distance left} metric below.
    \item {\bf Dev.}: deviation from the optimal classic path, i.e. A*.
    \item {\bf Comp.}: computation time in seconds.
    \item {\bf Dist. Left}: distance left to goal when failed, either because there is no path or the algorithm was not successful in finding a path.
    \item {\bf Search}: session search space, which is cells visited divided by total cells in a given map. 
\end{enumerate}

We report the averaged metrics for each test set. Another metric, map occupancy, was used to estimate the extent to which maps are occupied by obstacles. Table~\ref{tab:mapoccupancy} shows the occupancy rate of each map size. Results was computed using an Intel i7-6500u w/4 cores, 12GB, and an Nvidia GeForce 940M. Table~\ref{tab:specific} presents detailed comparative results.
Based on the results, WPN, WPN-map, WPN-view, and A* are able to find a path if available. Note that the reason even A* does not have 100\% success rate is that for some of the start/goal points, no path exists. According to Table~\ref{tab:mapoccupancy}, 8$\times$8 maps have the highest occupancy percentage, which means for randomly selected start/goal points, it is more likely that a path does not exist. This is why the success rate of 8$\times$8 is less than the other maps, even for A*. 
On small maps, i.e. 8$\times$8 and 
16$\times$16, the deviation of WPN from the optimal path is the least. On large maps, the deviation of WPN-map is less than the deviation of WPN. 
Additionally, when these algorithms fail, for the start/goal pairs with no paths, almost in all cases, WPN's \emph{Dist. Left} metric is the smallest. 

In terms of the computation time, none of the WPN variations are comparable to A*; however, all WPNs have smaller search space; which makes WPNs suitable for devices with low capabilities for search space, e.g. low memory. \textcolor{black}{Additionally, it is important to note, that while A* runs on CPU memory, WPN will run the global kernel on GPU memory. Combining this with the reduced search space allows for decreased load on the system's CPU and memory.} In summary, compared with A*, WPNs have smaller search space at the cost of being near-optimal. All models are available on the website of the project.

Compared with VIN and VIN64, WPN is consistently superior in success rate and deviation. Moreover; the performance of VIN degrades significantly as the size of the maps grow. Note that VIN and VIN64 are based on reinforcement learning and the search space for them is not applicable.

Finally, MPNet has better computation time on the benchmark maps compared with WPN, but is unable to achieve high success rates. To have a fair comparison, MPNet was retrained with the benchmark maps like other algorithms.

\begin{figure*}[]
    \centering
    \includegraphics[width=\textwidth]{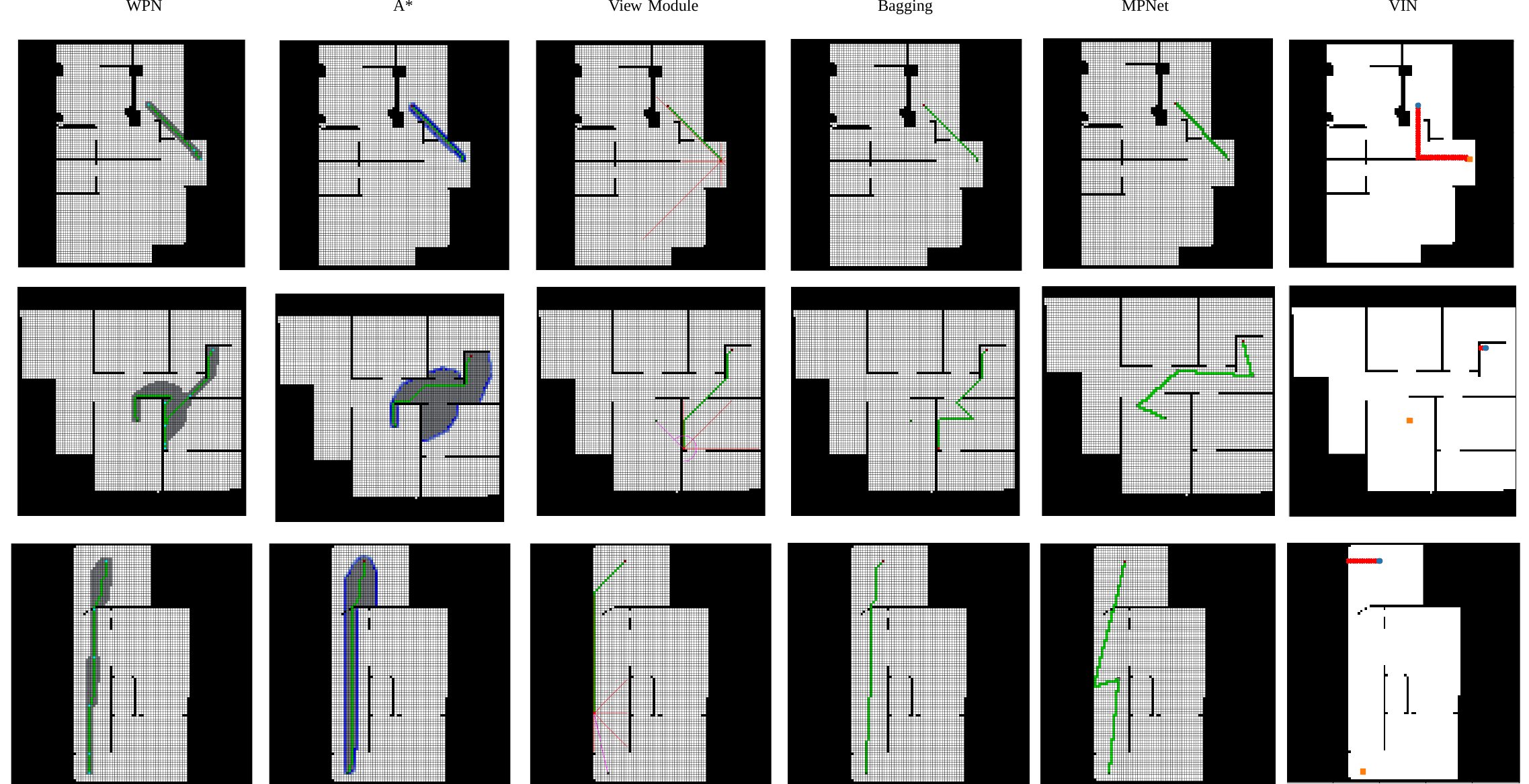}
    \caption{Results from HouseExpo~\cite{houseexpo}. WPN, A*, View Module, Bagging Module, MPNet, and VIN were run on three different HouseExpo maps. The search space of the WPN and A* algorithms can be seen in gray.}
    \vspace{-5 mm}
      \label{fig:he_maps}  
\end{figure*}

\subsection{Ablation Study} \label{sec:ablation}
\textcolor{black}{To justify WPN's architecture and its high success rate, we compare WPN against it's own components: view module, map module and bagging module. 
These individual modules of WPN were also trained and benchmarked the same way that WPN was trained. This is as an ablation study, and the results show that the individual modules are not able to achieve the performance of WPN, particularly in terms of the success rate. Note that the deviation from the optimal path is computed for the successful cases only, and this explains the better deviation percentage for those modules. 
The results can be seen in Table~\ref{tab:specific}. Without the waypoint module the success rates are not competitive to A*, as they are with WPN when including  waypoint module. This is predicted, as the local kernel planner is more successful in the smaller, known environments. Whereas the actual generation of the waypoints is done through the learned global kernel. The ability to generate learned waypoints and have a local kernel plan between them is what accounts for WPN's high success rates. }

\subsection{HouseExpo Benchmarking: Simulated Real-world Maps}

\begin{figure}
\begin{minipage}[h]{0.01\columnwidth}
  \centering
    \caption*{\rotatebox{90}{\quad \quad \textcolor{white}{--------}WPN}}
  \end{minipage}
  \hfill
  \begin{minipage}[b]{0.3\columnwidth}
  \centering
    \vspace{-2 mm}
    \includegraphics[width=.95\columnwidth]{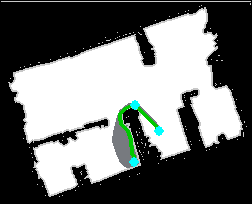}
  \end{minipage}
  \hfill
  \begin{minipage}[b]{0.3\columnwidth}
  \centering
    \includegraphics[width=.95\columnwidth]{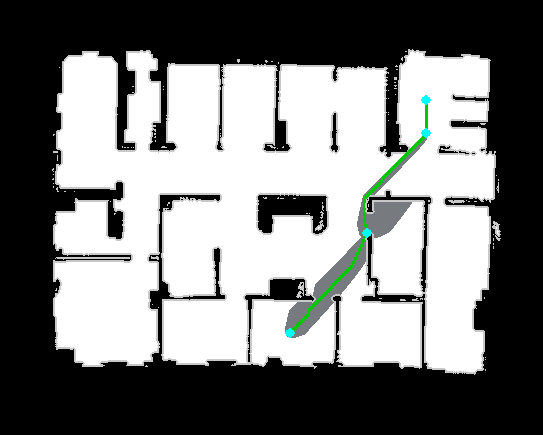}
  \end{minipage}
  \hfill
  \begin{minipage}[b]{0.3\columnwidth}
  \centering
    \includegraphics[width=\columnwidth]{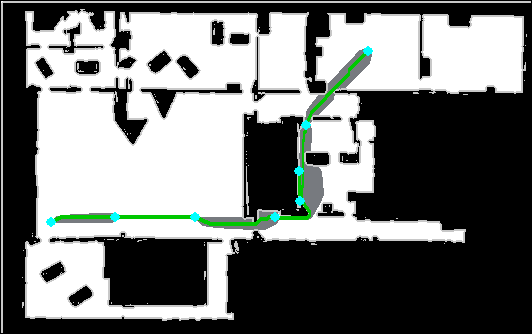}
  \end{minipage}
%%%
\vspace{-8 mm}
\newline
\begin{minipage}[b]{0.01\columnwidth}
  \centering
    \caption*{\rotatebox{90}{\quad \,\,  A*}}
  \end{minipage}
  \hfill
  \begin{minipage}[b]{0.3\columnwidth}
  \centering
    \vspace{-2 mm}
    \hspace{-2 mm}
    \includegraphics[width=.95\columnwidth]{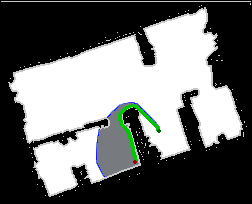}
  \end{minipage}
  \hfill
  \begin{minipage}[b]{0.3\columnwidth}
  \centering
      \hspace{-3 mm}
    \includegraphics[width=.95\columnwidth]{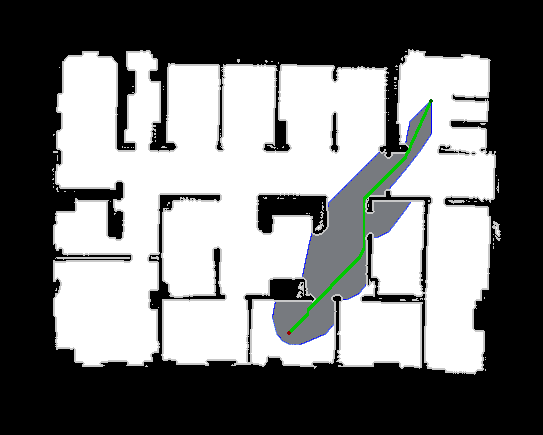}
  \end{minipage}
  \hfill
  \begin{minipage}[b]{0.3\columnwidth}
  \centering
        \hspace{-3 mm}
    \includegraphics[width=\columnwidth]{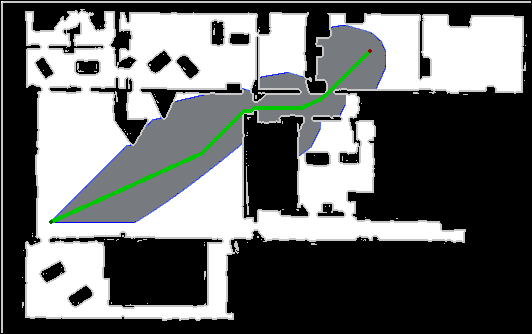}
  \end{minipage}
\caption{WPN (first row) vs A* (second row) in real-world maps, \cite{first_map}, \cite{second_map}, \cite{third_map}. Gray cells show the search space.}
  \label{fig:realmaps}  
  \vspace{-2 mm}
  \end{figure}

This experiment is designed to demonstrate the transferablity of the learned algorithm from synthetic maps to real-world maps. Here we use HouseExpo dataset~\cite{houseexpo}, which is a large-scale 2D floor plan dataset built on SUNCG~\cite{suncg}  dataset. We also test WPN on occupancy grid maps produced by SLAM. The maps that have been used are the following works: \cite{first_map}, \cite{second_map} and \cite{third_map}. 

\noindent{\it {\,\, 1) HouseExpo.}} From the HouseExpo dataset, 30 random maps were downsized to 100$\times$100 sizes. The maps were zero padded to make them square, then they were converted to PathBench acceptable format using the PathBench generator. All algorithms have higher success rates in HouseExpo, since these maps, based on Table~\ref{tab:mapoccupancy}, have low occupancy, and therefore it is more likely to find a path.\par

The algorithms used the same training models as Section~\ref{sec:Generalized}. Results can be seen in Table~\ref{tab:houseexpo}. \textcolor{black}{Results were computed on Intel Silver 4216 Cascade Lake, 32 cores w/ 128GB RAM and an
Nvidia V100 Volta GPU} We can see the results of WPN over all the other algorithms, where it surpasses in success rate and search space. WPN-map which uses the map module as its global kernel has the least deviation metric in indoor maps. Three HouseExpo maps with their paths are shown in Fig.~\ref{fig:he_maps}. All algorithms were not shown due to space limitation.

\begin{table}[t]
\centering
\caption{Percentage occupied by obstacles for the maps.}
\vspace{-2 mm}
\begin{tabular}{@{}c|c|c|c|c|c@{}}
\toprule
%\hline
\textbf{\scriptsize Map}                      & \textbf{\scriptsize 8$\times$8}               & \textbf{\scriptsize 16$\times$16}            & \textbf{\scriptsize 28$\times$28}             & \textbf{\scriptsize 64$\times$64}             & \textbf{\scriptsize HouseExpo}         \\ \midrule
{\textbf{\scriptsize Map Occupancy (\%) }} & {\scriptsize 23.6} & {\scriptsize 17.6} & {\scriptsize 16.1} & {\scriptsize 13.9} & {\scriptsize 3.58} \\ 
\bottomrule
%\hline
\end{tabular}
\label{tab:mapoccupancy}
\end{table}

\begin{table}[]
    \centering
    \caption{Generalization results of the algorithms on 30 100$\times$100 HouseExpo~\cite{houseexpo}.}
            \vspace{-2 mm}
            %\hline
            
    \begin{tabular}{@{}c|c|c|c|c|c@{}}
                \toprule
       %\hline
       \textbf{\scriptsize Planner} & \makecell {\textbf{\scriptsize Succ. } \\ \textbf{\scriptsize Rate (\%)} } & \makecell {\textbf{\scriptsize Dev.} \\ \textbf{\scriptsize }{\scriptsize (\%)} } & \makecell {\textbf{\scriptsize Comp.} \\ (sec) } &  \makecell{\textbf{\scriptsize Dist. Left} \\ \scriptsize{(when failed)} } & \makecell {\textbf{\scriptsize Search} \\ \textbf{\scriptsize (\%)} }  \\
       \hline
       %\hline
       {\scriptsize WPN}              & {\bf 100} & -29.6 & 8.235 & 0 & 1.12 \\
       %\hline
        {\scriptsize WPN-view}             & {\bf 100} & -38.5 & 1.5476 & 0 & {\bf 1.07} \\
       %\hline
        {\scriptsize WPN-map}              & {\bf 100} & {\bf -13.96} & 1.4708 & 0 & 1.28 \\
       %\hline
       \hline
       {\scriptsize A*}               & {\bf 100} & 0 & {\bf 0.2414} & 0 & 6.12 \\
       \hline
       %\hline
       {\scriptsize Bagging module}   & 90.0 & -1.89 & 4.364 & 1.06 & NA \\
       %\hline
       {\scriptsize View module \cite{nicola2018lstm}} & 76.60 & -1.99  & 0.3948 & 5.72 & NA \\
       %\hline
       {\scriptsize Map module \cite{inoue2019robot}} & 80 & -3.54 & 0.2932 & 5.85 & NA \\
       \hline
       {\scriptsize VIN \cite{tamar2016value}} & 51.72 & -50.2 & 0.349 & 50.8 & NA \\
       \hline
       {\scriptsize MPNet \cite{qureshi2019motion}} & 83.33 & -42.3&  0.658& 64.8 & NA \\
       %\hline
       \bottomrule
    \end{tabular}
    %\hline
    \label{tab:houseexpo}
        \vspace{-2 mm}
\end{table}

\noindent{\it {\,\, 2) Occupancy Grid Maps.}} WPN was trained only on synthetic images. Fig.
\ref{fig:realmaps} highlights the performance of the WPN against A* on the real-world occupancy grid maps. The gray cells demonstrate the search space of both algorithms. We can notice that the algorithm maintains the same behaviour across different environments which confirms the robustness to unknown environments. This is intuitively correct, as we have used machine learning methods to find the path, and thus, we inherit the generalization properties.

\begin{figure}
  \begin{minipage}[b]{0.115\textwidth}
  \centering
    \includegraphics[width=\linewidth]{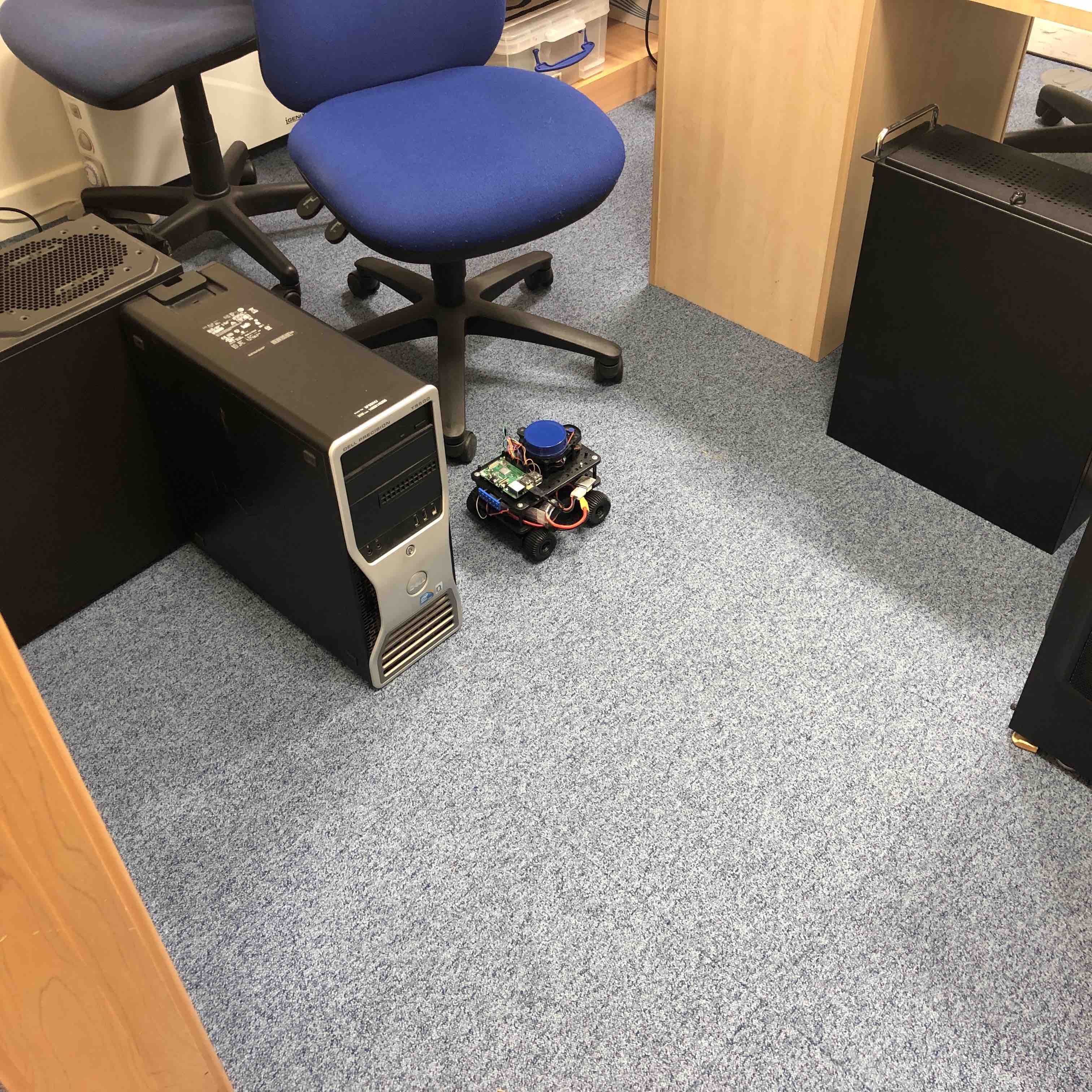}
  \end{minipage}
  \hfill
  \begin{minipage}[b]{0.115\textwidth}
  \centering
    \includegraphics[width=\linewidth]{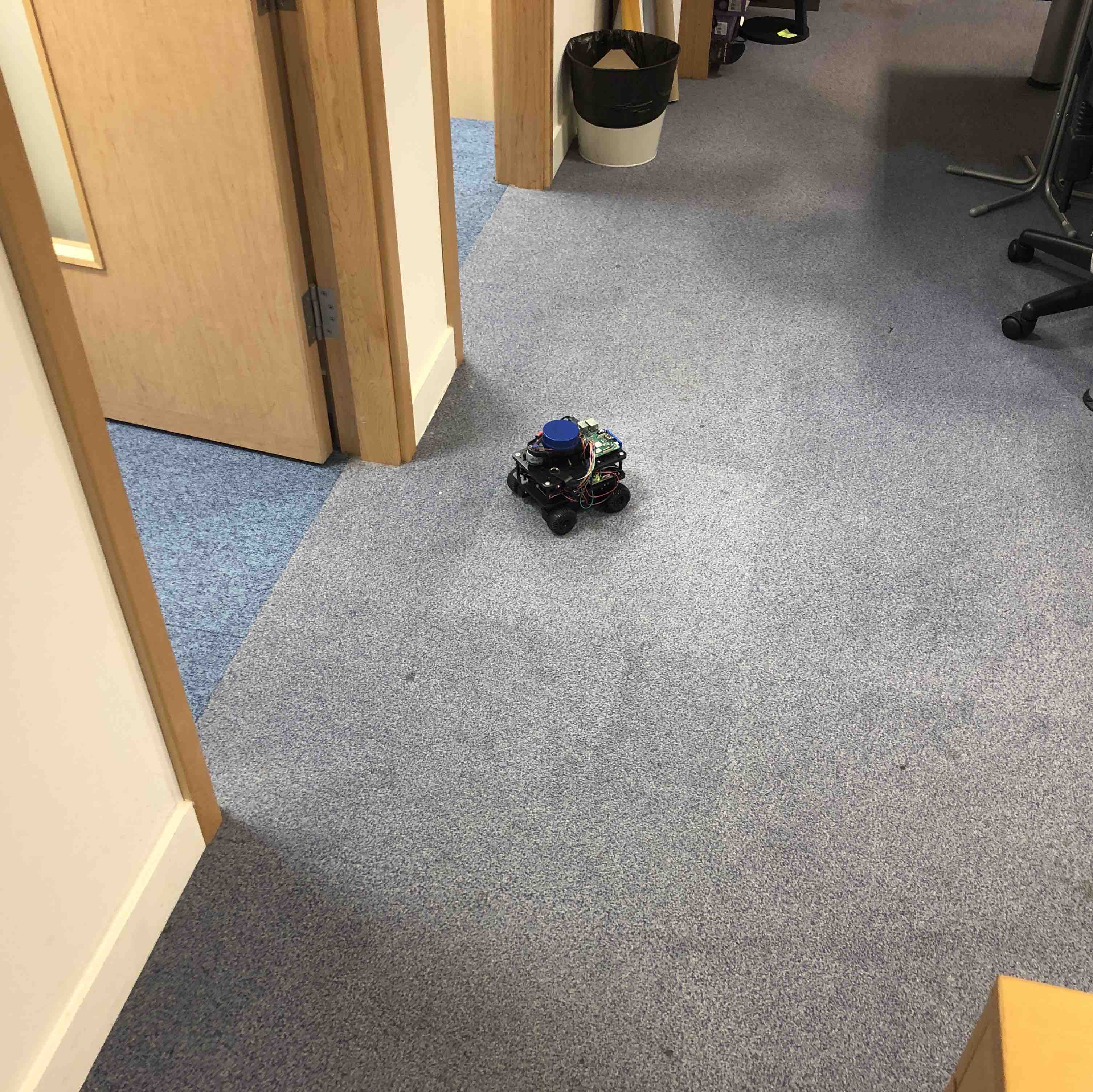}
  \end{minipage}
  \hfill\
  \begin{minipage}[b]{0.115\textwidth}
  \centering
    \includegraphics[width=\linewidth]{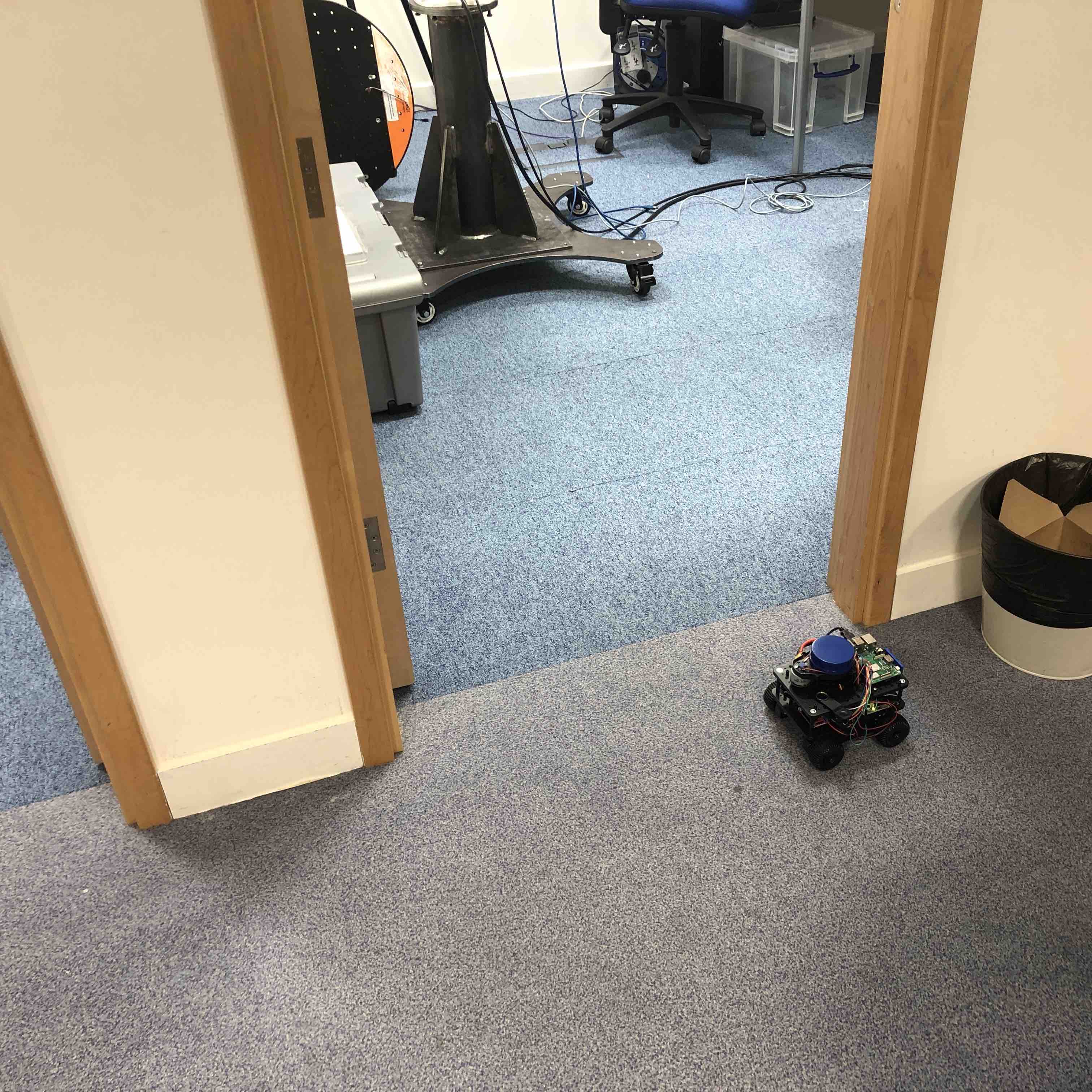}
  \end{minipage}
  \hfill
  \begin{minipage}[b]{0.115\textwidth}
  \centering
    \includegraphics[width=\linewidth]{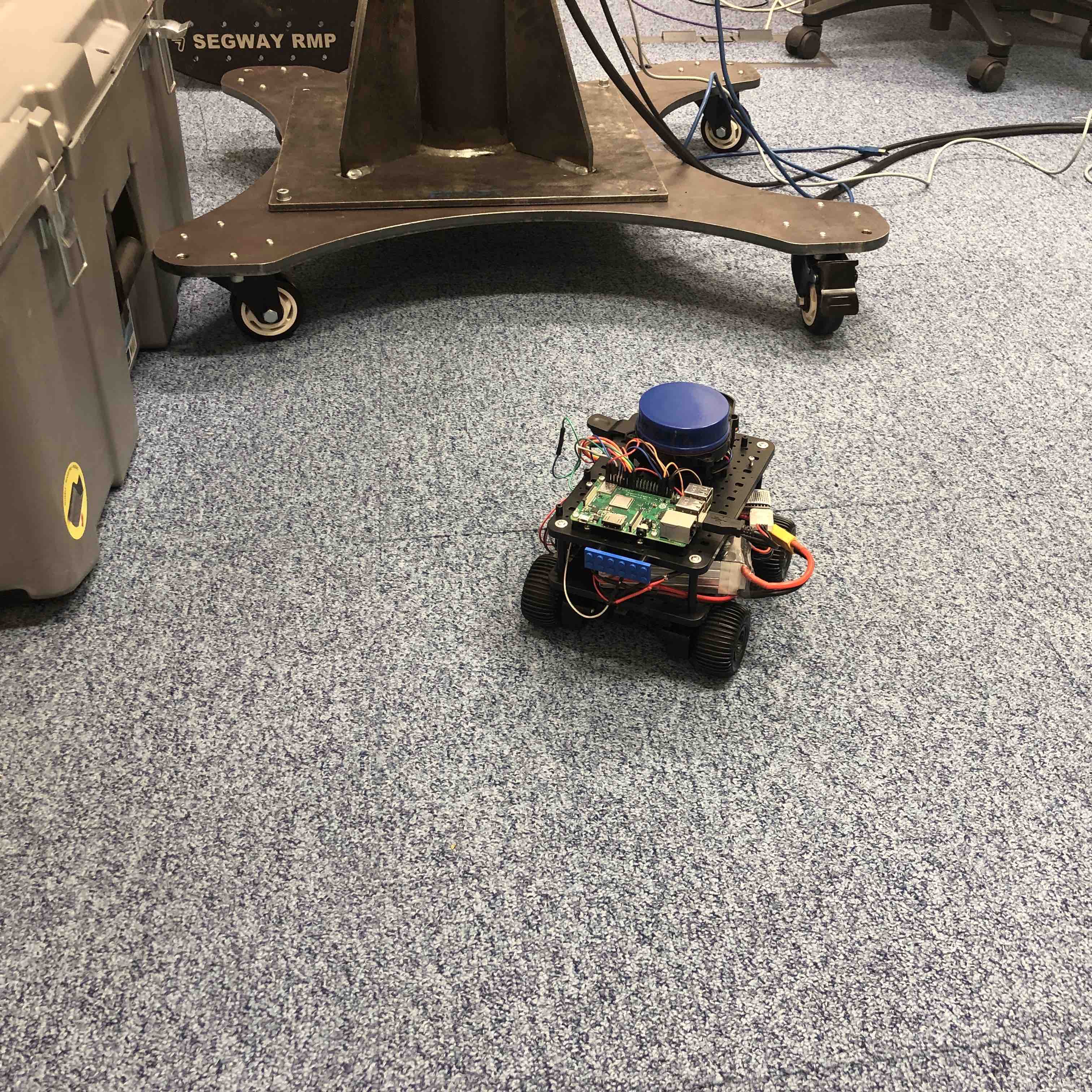}
    \end{minipage}  
    %%%%%%%%%%%%%%%%%
  \begin{minipage}[b]{0.115\textwidth}
  \centering
    \includegraphics[width=\linewidth]{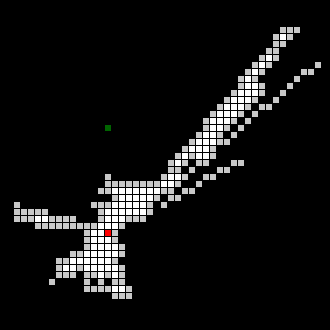}
  \end{minipage}
  \hfill
  \begin{minipage}[b]{0.115\textwidth}
  \centering
    \includegraphics[width=\linewidth]{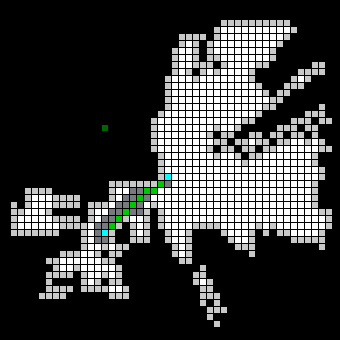}
  \end{minipage}
  \hfill
  \begin{minipage}[b]{0.115\textwidth}
  \centering
    \includegraphics[width=\linewidth]{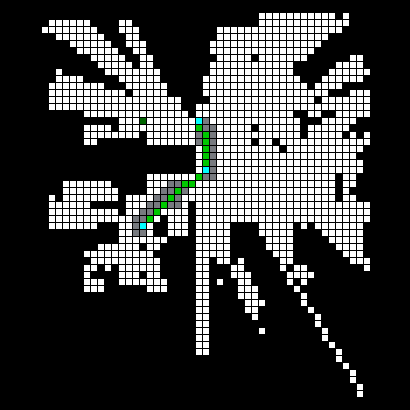}
  \end{minipage}
  \hfill
  \begin{minipage}[b]{0.115\textwidth}
  \centering
    \includegraphics[width=\linewidth]{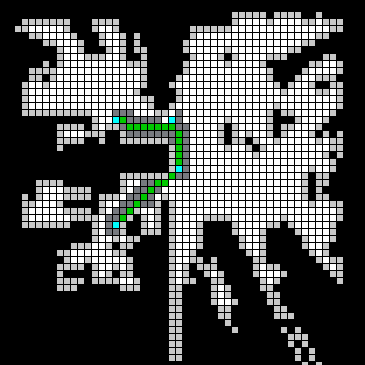}    
    \end{minipage}  
%%%
  \caption{Real-world robot navigation using WPN via ROS. 
  The robot is asked to move to a target in the adjacent room, while there is no global map available. Top row represents real-world view of the robot, the second row represents the live map, as being updated (left to right shows the progress through time). The true dimension of the grid is 128$\times$128.}
  \vspace{-1 mm}
  \label{fig: robot_run}
  \end{figure}

\begin{figure}
\begin{minipage}{.48\columnwidth}
  \centering
  \includegraphics[width=\linewidth]{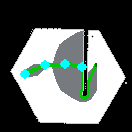}
  \caption*{WPN-map (len. = 134.77m)}
  \label{fig:test1}
\end{minipage}%
\hspace{1pt}
\begin{minipage}{.48\columnwidth}
  \centering
  \includegraphics[width=\linewidth]{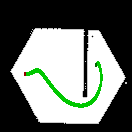}
  \caption*{A* (len. = 163.90m)}
  \label{fig:test2}
\end{minipage}
\vspace{-2 mm}
\caption{WPN-map vs A* on the live exploration experiment in an unknown Gazebo world. A* uses the frontier exploration algorithm and subsequently plan to the goal. WPN-map is able to plan and explore simultaneously. The map size is 128$\times$128 and Turtlebot3 with a scanning laser ranger is used. }
  \vspace{-3 mm}
  \label{fig:ros-exploration}  
  \end{figure}
\vspace{-1 mm}

\begin{figure}[t]
\centering
    \includegraphics[width=.45\columnwidth]{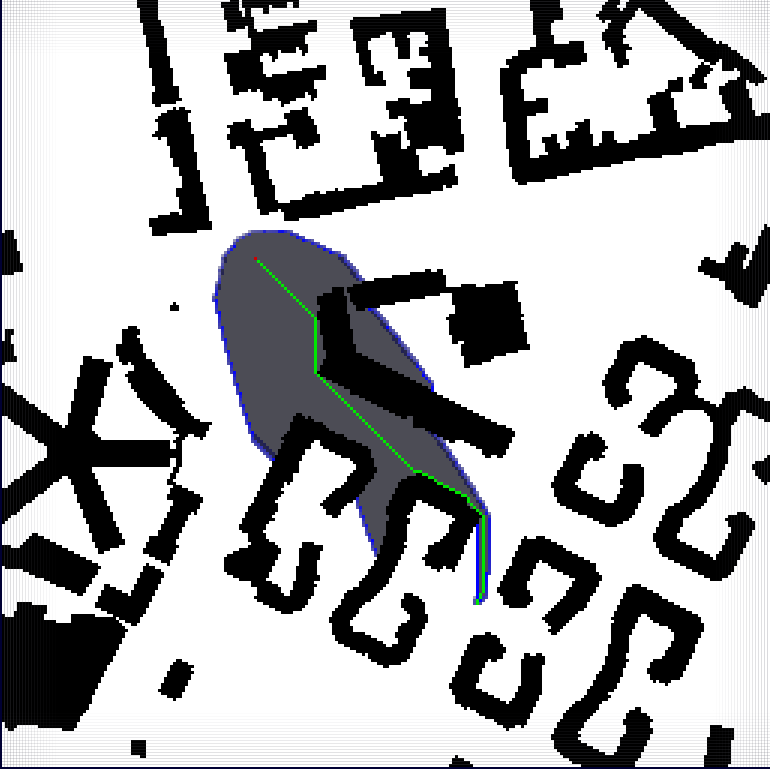}
    \includegraphics[width=.49\columnwidth]{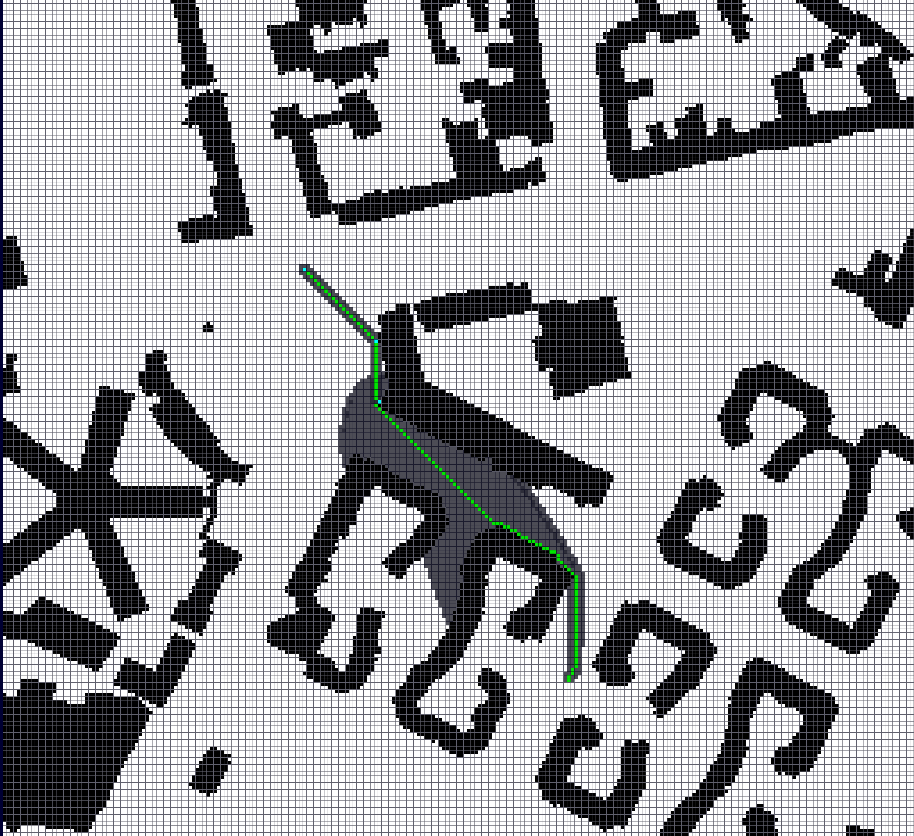}

  \vspace{1mm}
\caption{City maps (512$\times$512) used for large map experimentation. A* path on the left, and WPN-view on the right}
  \label{fig:largemaps}  
  \vspace{-2 mm}
  \end{figure}
\subsection{Real-world and Gazebo Experiments}
Several real-world experiment were done with a differential drive robot, one of which reported here, the rest are in the video. The robot has a YDLidar sensor, a 360-degree two-dimensional scanning laser ranger, used to generate a 2D map using the GMapping algorithm~\cite{gmapping}. The motherboard of the robot is a Raspberry Pi board which is running Raspbian, and makes use of ROS. The robot is asked to move from one room to another, without having a global map. This is a task that A* is not able to find a path for. Using WPN, the algorithm suggests a waypoint and a path to it, then the robot navigates to the waypoint. While the map gets updated, WPN updates the waypoints to guide the robot to the final goal. Fig.~\ref{fig: robot_run} showcases the robot.

We also run another experiment in Gazebo to further demonstrate the benefit of WPN over A*. In this experiment, the frontier exploration algoritm~\cite{Yamauchi} has been added to the A* algorithm. A simple world was generated in Gazebo, with a start and goal point chosen. The map is unknown and planning was done with WPN-map and A* with frontier exploration, on a Turtlebot3 robot with a scanning laser ranger. A* must explore and subsequently plan. WPN is able to simultaneously plan and explore, which leads to a lower overall path length compared to A* with exploration. This can be seen in Figure~\ref{fig:ros-exploration}. The start point is on the left side, the goal is on the right side. 
\subsection{Large Maps}
\textcolor{black}{To demonstrate the performance of WPN in larger map environments, we test the algorithm on 30 large 2D environments, and compare them to A*.}
Real world city maps~\cite{sturtevant2012benchmarks} are used for testing. These maps are 512$\times$512. WPN was trained on 45,000 64$\times$64 maps, split between types uniform random fill, block map, and house maps. Computation was done on an Intel Silver 4216 Cascade Lake CPU (16 cores) with 64GB memory, and an Nvidia V100 GPU. Results can be seen in Table~\ref{tab:largemaps}. Sample runs can be seen in Figure~\ref{fig:largemaps}. \textcolor{black}{It can be seen that WPN and it's variants are able to compete with A* in terms of success rate, while maintaining a significantly lower search space. We also compare other ML approaches, all of which perform rather poorly on these large scale environments. We can see that WPN performs very well on large scale maps. The success rate can be attributed to the waypoint module, as discussed in Section~\ref{sec:ablation}}

\begin{table}[] 
\caption{Results from large map experiment, demonstrating WPN's ability to plan on large maps while maintaining a high accuracy and low search space. Maps used were city style maps~\cite{sturtevant2012benchmarks}. Algorithms were run on 30 maps, with two attempts per map (with different start and goal points for each attempt.)}
\centering
\resizebox{\columnwidth}{!}{
\begin{tabular}{@{}cccc@{}}
\toprule
                             & Accuracy (\%) & Deviation (\%) & Search Space (\%) \\ \midrule
\multicolumn{1}{c|}{A*}      & 100     & 0        & 7.22         \\

\multicolumn{1}{c|}{WPN} & 100        &   8.47      & 3.23        \\
\multicolumn{1}{c|}{WPN-map}& 100        & 6.90  & 3.37      \\ 
\multicolumn{1}{c|}{WPN-view}& 100        & 5.31 & 4.19     \\ 
\multicolumn{1}{c|}{View Module}& 16.7        & 37.0  & NA     \\ 
\multicolumn{1}{c|}{Map Module}& 35.0        & 2.51 & NA     \\

\bottomrule
\end{tabular}
}
\label{tab:largemaps}
\end{table}

%\vspace{-1 mm}
\section{Conclusion}
%\vspace{-1 mm}
\label{sec:conclusion}
In this paper a novel learning-based path planning algorithm was proposed. The algorithm, waypoint planning networks (WPN), was trained with synthetic maps and tested on various types of maps and a real robot. WPN presents a significant advantage over A* in terms of search space, while achieving the success rate of A*. WPN has the benefit of working with partial maps, while also maintaining a high efficiency and low deviation from A*. It also provides a significant advantage over other learning-based algorithms in terms of success rate and deviation from optimal paths. WPN also generalizes more successfully than other learned algorithms and is easily trained. \textcolor{black}{It is able to compete with A* on large scale environments, where other learned approaches perform poorly.} It takes advantage of the benefits of both learned algorithms and classic algorithms. 

In the future, WPN will be optimized to use a different global kernel, such as generative adversarial networks. WPN will be trained on different real-world datasets, to generate better results. We also plan to improve multiprocessing on WPN, which will significantly improve computation time. Moreover, we plan to extend WPN to 3D maps and higher dimensions to perform path planning for higher degrees-of-freedom and complex robotic systems such as manipulators. \textcolor{black}{Also, we will include kinematics constraints in the architecture to account for real-world physical constraints.}

\section*{Acknowledgment}
This work was partially funded by DRDC-IDEaS (CPCA-0126). 
We gratefully acknowledge the support of NVIDIA Corporation with the donation of the Titan Xp GPU used for conducting experiments.

\bibliographystyle{IEEEtran}
\bibliography{main}

\begin{thebibliography}{10}
\providecommand{\url}[1]{#1}
\csname url@samestyle\endcsname
\providecommand{\newblock}{\relax}
\providecommand{\bibinfo}[2]{#2}
\providecommand{\BIBentrySTDinterwordspacing}{\spaceskip=0pt\relax}
\providecommand{\BIBentryALTinterwordstretchfactor}{4}
\providecommand{\BIBentryALTinterwordspacing}{\spaceskip=\fontdimen2\font plus
\BIBentryALTinterwordstretchfactor\fontdimen3\font minus
  \fontdimen4\font\relax}
\providecommand{\BIBforeignlanguage}[2]{{%
\expandafter\ifx\csname l@#1\endcsname\relax
\typeout{** WARNING: IEEEtran.bst: No hyphenation pattern has been}%
\typeout{** loaded for the language `#1'. Using the pattern for}%
\typeout{** the default language instead.}%
\else
\language=\csname l@#1\endcsname
\fi
#2}}
\providecommand{\BIBdecl}{\relax}
\BIBdecl

\bibitem{gonzalez2016review}
D.~Gonz{\'a}lez, J.~P{\'e}rez, V.~Milan{\'e}s, and F.~Nashashibi, ``{A}
  {R}eview of {M}otion {P}lanning {T}echniques for {A}utomated {V}ehicles.''
  \emph{IEEE Trans. Intelligent Transportation Systems}, vol.~17, no.~4, pp.
  1135--1145, 2016.

\bibitem{choset2005principles}
H.~M. Choset, S.~Hutchinson, K.~M. Lynch, G.~Kantor, W.~Burgard, L.~E. Kavraki,
  and S.~Thrun, \emph{{Principles of Robot Motion: theory, algorithms, and
  implementation}}.\hskip 1em plus 0.5em minus 0.4em\relax MIT press, 2005.

\bibitem{duchovn2014path}
F.~Ducho{\v{n}}, A.~Babinec, M.~Kajan, P.~Be{\v{n}}o, M.~Florek, T.~Fico, and
  L.~Juri{\v{s}}ica, ``{Path Planning with Modified {A} Star Algorithm for a
  Mobile Robot},'' \emph{Procedia Engineering}, vol.~96, pp. 59--69, 2014.

\bibitem{zhang2014multiple}
Z.~Zhang and Z.~Zhao, ``{A Multiple Mobile Robots Path Planning Algorithm based
  on {A}-star and {Dijkstra} Algorithm},'' \emph{International Journal of Smart
  Home}, vol.~8, no.~3, pp. 75--86, 2014.

\bibitem{5937169}
W.~Y. {Loong}, L.~Z. {Long}, and L.~C. {Hun}, ``{A Star Path Following Mobile
  Robot},'' in \emph{International Conference on Mechatronics (ICOM)}, 2011,
  pp. 1--7.

\bibitem{lavalle1998rapidly}
S.~M. LaValle, ``{Rapidly-exploring Random Trees: A new tool for path
  planning},'' 1998.

\bibitem{rodriguez2006obstacle}
S.~Rodriguez, X.~Tang, J.-M. Lien, and N.~M. Amato, ``{A}n obstacle-based
  rapidly-exploring random tree,'' in \emph{International Conference on
  Robotics and Automation}, 2006, pp. 895--900.

\bibitem{lavalle2001randomized}
S.~M. LaValle and J.~J. Kuffner~Jr, ``{R}andomized kinodynamic planning,''
  \emph{The international journal of robotics research}, vol.~20, no.~5, pp.
  378--400, 2001.

\bibitem{karaman2011sampling}
S.~Karaman and E.~Frazzoli, ``{S}ampling-based algorithms for optimal motion
  planning,'' \emph{The international journal of robotics research}, vol.~30,
  no.~7, pp. 846--894, 2011.

\bibitem{szepesvari2010algorithms}
C.~Szepesv{\'a}ri, ``{A}lgorithms for reinforcement learning,'' \emph{Synthesis
  lectures on artificial intelligence and machine learning}, vol.~4, no.~1, pp.
  1--103, 2010.

\bibitem{satia1973markovian}
J.~K. Satia and R.~E. Lave~Jr, ``Markovian decision processes with uncertain
  transition probabilities,'' \emph{Operations Research}, vol.~21, no.~3, pp.
  728--740, 1973.

\bibitem{pathbench}
{Alexandru-Iosif Toma, Hao-Ya Hsueh, Hussein Ali Jaafar, Stephen James, Daniel
  Lenton, Ronald Clark, Sajad Saeedi}, ``{PathBench}: {A} {Benchmarking}
  {Platform} for {Classical} and {Learned} {Path} {Planning} {Algorithms},'' in
  \emph{IEEE Conference on Robotics and Vision, Burnaby BC Canada}, 2021.

\bibitem{tamar2016value}
A.~Tamar, Y.~Wu, G.~Thomas, S.~Levine, and P.~Abbeel, ``{V}alue {I}teration
  {N}etworks,'' in \emph{Advances in Neural Information Processing Systems},
  2016, pp. 2154--2162.

\bibitem{qureshi2019motion}
A.~H. Qureshi, A.~Simeonov, M.~J. Bency, and M.~C. Yip, ``Motion planning
  networks,'' in \emph{2019 International Conference on Robotics and Automation
  (ICRA)}.\hskip 1em plus 0.5em minus 0.4em\relax IEEE, 2019, pp. 2118--2124.

\bibitem{luo2014effective}
C.~Luo, M.~Krishnan, M.~Paulik, and G.~E. Jan, ``{An Effective Trace-guided
  Wavefront Navigation and Map-building Approach for Autonomous Mobile
  Robots},'' in \emph{Intelligent Robots and Computer Vision}, vol. 9025, 2014,
  p. 90250U.

\bibitem{rajko2001pursuit}
S.~Rajko and S.~M. LaValle, ``A pursuit-evasion bug algorithm,'' in \emph{IEEE
  International Conference on Robotics and Automation (ICRA)}, vol.~2, 2001,
  pp. 1954--1960.

\bibitem{kavraki1994probabilistic}
L.~Kavraki, P.~Svestka, and M.~H. Overmars, \emph{{Probabilistic Roadmaps for
  Path Planning in High-dimensional Configuration Spaces}}, 1994, vol. 1994.

\bibitem{reeds1990optimal}
J.~Reeds and L.~Shepp, ``{Optimal Paths for a Car that Goes both Forwards and
  Backwards},'' \emph{Pacific journal of mathematics}, vol. 145, no.~2, pp.
  367--393, 1990.

\bibitem{funke2012up}
J.~Funke, P.~Theodosis, R.~Hindiyeh, G.~Stanek, K.~Kritatakirana, C.~Gerdes,
  D.~Langer, M.~Hernandez, B.~M{\"u}ller-Bessler, and B.~Huhnke, ``{Up to the
  Limits: Autonomous {Audi TTS}},'' in \emph{IEEE Intelligent Vehicles
  Symposium}, 2012, pp. 541--547.

\bibitem{xu2012real}
W.~Xu, J.~Wei, J.~M. Dolan, H.~Zhao, and H.~Zha, ``{A Real-time Motion Planner
  with Trajectory Optimization for Autonomous Vehicles},'' in \emph{IEEE
  International Conference on Robotics and Automation (ICRA)}, 2012, pp.
  2061--2067.

\bibitem{dolgov2010path}
D.~Dolgov, S.~Thrun, M.~Montemerlo, and J.~Diebel, ``{Path Planning for
  Autonomous Vehicles in Unknown Semi-structured Environments},'' \emph{The
  International Journal of Robotics Research}, vol.~29, no.~5, pp. 485--501,
  2010.

\bibitem{ziegler2014making}
J.~Ziegler, P.~Bender, M.~Schreiber, H.~Lategahn, T.~Strauss, C.~Stiller,
  T.~Dang, U.~Franke, N.~Appenrodt, C.~G. Keller \emph{et~al.}, ``{Making
  {Bertha} Drive—An Autonomous Journey on a Historic Route},'' \emph{IEEE
  Intelligent transportation systems magazine}, vol.~6, no.~2, pp. 8--20, 2014.

\bibitem{Chen2016Humanoids}
N.~{Chen}, M.~{Karl}, and P.~{van der Smagt}, ``{Dynamic Movement Primitives in
  Latent Space of Time-dependent Variational Autoencoders},'' in \emph{IEEE-RAS
  International Conference on Humanoid Robots (Humanoids)}, 2016, pp. 629--636.

\bibitem{gupta2017cognitive}
S.~Gupta, J.~Davidson, S.~Levine, R.~Sukthankar, and J.~Malik, ``{Cognitive
  mapping and planning for visual navigation},'' in \emph{IEEE Conference on
  Computer Vision and Pattern Recognition (CVPR)}, 2017, pp. 2616--2625.

\bibitem{inoue2019robot}
M.~Inoue, T.~Yamashita, and T.~Nishida, ``{R}obot {P}ath {P}lanning by {LSTM}
  {N}etwork {U}nder {C}hanging {E}nvironment,'' in \emph{Advances in Computer
  Communication and Computational Sciences}.\hskip 1em plus 0.5em minus
  0.4em\relax Springer, 2019, pp. 317--329.

\bibitem{NeuralRRT}
J.~{Wang}, W.~{Chi}, C.~{Li}, C.~{Wang}, and M.~Q. .~H. {Meng}, ``{Neural
  RRT*}: Learning-based optimal path planning,'' \emph{IEEE Transactions on
  Automation Science and Engineering}, vol.~17, no.~4, pp. 1748--1758, 2020.

\bibitem{ross2011reduction}
S.~Ross, G.~Gordon, and D.~Bagnell, ``A reduction of imitation learning and
  structured prediction to no-regret online learning,'' in \emph{Proceedings of
  the fourteenth international conference on artificial intelligence and
  statistics}, 2011, pp. 627--635.

\bibitem{TDPPNet}
K.~Wu, M.~{Abolfazli Esfahani}, S.~Yuan, and H.~Wang, ``{TDPP-Net}: Achieving
  three-dimensional path planning via a deep neural network architecture,''
  \emph{Neurocomputing}, vol. 357, pp. 151 -- 162, 2019.

\bibitem{qureshi2018deeply}
A.~H. Qureshi and M.~C. Yip, ``{Deeply Informed Neural Sampling for Robot
  Motion Planning},'' in \emph{IEEE/RSJ International Conference on Intelligent
  Robots and Systems (IROS)}, 2018, pp. 6582--6588.

\bibitem{chamzas2019using}
C.~Chamzas, A.~Shrivastava, and L.~E. Kavraki, ``{Using Local Experiences for
  Global Motion Planning},'' \emph{arXiv preprint arXiv:1903.08693}, 2019.

\bibitem{qureshi2020motion}
A.~H. Qureshi, Y.~Miao, A.~Simeonov, and M.~C. Yip, ``Motion planning networks:
  Bridging the gap between learning-based and classical motion planners,''
  \emph{IEEE Transactions on Robotics}, pp. 1--9, 2020.

\bibitem{Bit*}
J.~D. {Gammell}, S.~S. {Srinivasa}, and T.~D. {Barfoot}, ``{Batch Informed
  Trees ({BIT}*): Sampling-based Optimal Planning via the Heuristically Guided
  Search of Implicit Random Geometric Graphs},'' in \emph{IEEE International
  Conference on Robotics and Automation (ICRA)}, 2015, pp. 3067--3074.

\bibitem{CoMPNet}
A.~H. {Qureshi}, J.~{Dong}, A.~{Choe}, and M.~C. {Yip}, ``Neural manipulation
  planning on constraint manifolds,'' \emph{IEEE Robotics and Automation
  Letters}, vol.~5, no.~4, pp. 6089--6096, 2020.

\bibitem{Johnson_IROS_2020}
J.~J. {Johnson}, L.~{Li}, F.~{Liu}, A.~H. {Qureshi}, and M.~C. {Yip},
  ``Dynamically constrained motion planning networks for non-holonomic
  robots,'' in \emph{IROS}, 2020, pp. 6937--6943.

\bibitem{bency2019neural}
M.~J. Bency, A.~H. Qureshi, and M.~C. Yip, ``{Neural Path Planning: Fixed Time,
  Near-Optimal Path Generation via Oracle Imitation},'' \emph{arXiv preprint
  arXiv:1904.11102}, 2019.

\bibitem{Levine2013}
S.~Levine and V.~Koltun, ``Guided policy search,'' in \emph{International
  Conference on International Conference on Machine Learning (ICML)}, 2013, pp.
  III--1--III--9.

\bibitem{Abbeel2010}
P.~Abbeel, A.~Coates, and A.~Y. Ng, ``{Autonomous Helicopter Aerobatics Through
  Apprenticeship Learning},'' \emph{International Journal of Robotics
  Research}, vol.~29, no.~13, pp. 1608--1639, 2010.

\bibitem{srinivas2018universal}
A.~Srinivas, A.~Jabri, P.~Abbeel, S.~Levine, and C.~Finn, ``Universal planning
  networks,'' \emph{International Conference in Machine Learning}, 2018.

\bibitem{nicola2018lstm}
F.~Nicola, Y.~Fujimoto, and R.~Oboe, ``A {LSTM} {N}eural {N}etwork applied to
  {M}obile {R}obots {P}ath {P}lanning,'' in \emph{IEEE International Conference
  on Industrial Informatics (INDIN)}, 2018, pp. 349--354.

\bibitem{first_map}
A.~Howard, L.~Parker, and G.~S.~Sukhatme, ``{The {SDR} Experience: Experiments
  with a Large-Scale Heterogeneous Mobile Robot Team},'' vol.~21, 2004.

\bibitem{second_map}
S.~Gholami~Shahbandi and M.~Magnusson, ``{{2D} Map Alignment with Region
  Decomposition},'' \emph{Autonomous Robots}, vol.~43, no.~5, pp. 1117--1136,
  2019.

\bibitem{third_map}
R.~Tang, ``{C}ustom {P}layer plugins,'' available:
  \url{http://robotang.co.nz/projects/robotics/custom-player-plugins/};
  accessed \today.

\bibitem{houseexpo}
T.~Li, D.~Ho, C.~Li, D.~Zhu, C.~Wang, and M.~Meng, ``{HouseExpo}: A large-scale
  {2D} indoor layout dataset for learning-based algorithms on mobile robots,''
  in \emph{IEEE/RSJ International Conference on Intelligent Robots and Systems
  (IROS)}, 2020.

\bibitem{hochreiter1997long}
S.~Hochreiter and J.~Schmidhuber, ``{L}ong short-term memory,'' \emph{Neural
  computation}, vol.~9, no.~8, pp. 1735--1780, 1997.

\bibitem{Everett_IROS_2019}
M.~{Everett}, J.~{Miller}, and J.~P. {How}, ``Planning beyond the sensing
  horizon using a learned context,'' in \emph{2019 IROS}, 2019, pp. 1064--1071.

\bibitem{dietterich2000ensemble}
T.~G. Dietterich, ``{Ensemble Methods in Machine Learning},'' in
  \emph{International workshop on multiple classifier systems}.\hskip 1em plus
  0.5em minus 0.4em\relax Springer, 2000, pp. 1--15.

\bibitem{suncg}
S.~{Song}, F.~{Yu}, A.~{Zeng}, A.~X. {Chang}, M.~{Savva}, and T.~{Funkhouser},
  ``Semantic scene completion from a single depth image,'' in \emph{2017 IEEE
  Conference on Computer Vision and Pattern Recognition {(CVPR)}}, 2017, pp.
  190--198.

\bibitem{gmapping}
G.~{Grisetti}, C.~{Stachniss}, and W.~{Burgard}, ``{Improved Techniques for
  Grid Mapping With {R}ao-{B}lackwellized Particle Filters},'' \emph{IEEE
  Transactions on Robotics}, vol.~23, no.~1, pp. 34--46, 2007.

\bibitem{Yamauchi}
B.~Yamauchi, A.~Schultz, and W.~Adams, ``Integrating exploration and
  localization for mobile robots,'' \emph{Adaptive Behavior}, vol.~7, no.~2,
  pp. 217--229, 1999.

\bibitem{sturtevant2012benchmarks}
N.~R. Sturtevant, ``{Benchmarks for Grid-based Pathfinding},'' \emph{IEEE
  Transactions on Computational Intelligence and AI in Games}, vol.~4, no.~2,
  pp. 144--148, 2012.

\end{thebibliography}

\end{document}